\documentclass{ifm-tr}

\usepackage{amsmath,amsfonts,bm}

\def\eqref#1{equation~\ref{#1}}

\def\1{\bm{1}}

\DeclareMathAlphabet{\mathsfit}{\encodingdefault}{\sfdefault}{m}{sl}
\SetMathAlphabet{\mathsfit}{bold}{\encodingdefault}{\sfdefault}{bx}{n}

\usepackage[utf8]{inputenc}
\usepackage[T1]{fontenc}
\usepackage{times}

\usepackage{amsmath}
\usepackage{amssymb}
\usepackage{amsfonts}
\usepackage{bbm}
\usepackage{nicefrac}
\usepackage{siunitx}
\usepackage{etoolbox}
\usepackage{caption}

\usepackage{booktabs}
\usepackage{multirow}
\usepackage{makecell}
\usepackage{hhline}
\usepackage{longtable}
\usepackage{nicematrix}
\usepackage{colortbl}
\usepackage{wrapfig}

\usepackage{graphicx}
\usepackage{epsfig}
\usepackage{wrapfig}
\usepackage{float}
\usepackage{floatflt}
\usepackage{subcaption}
\usepackage{tikz}

\usepackage{algorithm}
\usepackage{algorithmic}

\usepackage{xcolor}

\definecolor{lightred}{rgb}{1,0.8,0.8}
\definecolor{lightgreen}{rgb}{0.8,1,0.8}
\definecolor{lightblue}{rgb}{0.88,0.96,1}
\definecolor{lightgray}{rgb}{0.9,0.9,0.9}
\definecolor{darkgreen}{RGB}{0,100,0}
\definecolor{cellHighlight}{HTML}{dbefff}

\usepackage{enumitem}
\usepackage{xspace}
\usepackage{pifont}
\usepackage{fontawesome5}
\usepackage{soul}
\usepackage[normalem]{ulem} 
\usepackage{multicol}
\usepackage{comment}
\usepackage{lipsum}

\usepackage{xurl}
\usepackage{url}
\usepackage{hyperref}

\usepackage{microtype}

\usepackage{thmtools}

\usepackage[frozencache,cachedir=.]{minted}
\usepackage{tcolorbox}

\usepackage{eccvabbrv}

\titleformat{\section}{\large\sffamily\bfseries\color{black}}{\thesection}{1em}{}
\titleformat{\subsection}{\normalsize\sffamily\bfseries\color{black}}{\thesubsection}{1em}{}
\titleformat{\subsubsection}{\normalsize\sffamily\bfseries\color{black}}{\thesubsubsection}{1em}{}

\newcommand{\cmark}{\textcolor{green!60!black}{\ding{51}}}
\newcommand{\xmark}{\textcolor{red!75!black}{\ding{55}}}

\newcommand\rurl[1]{%
    \href{https://#1}{\nolinkurl{#1}}%
}

\title{WorldGuide: Goal-Directed Video World Model for\\[10pt]
Procedural Task Execution}

\author[]{\textbf{Ankan Deria, Komal Kumar, Hisham Cholakkal, Fahad Shahbaz Khan, Salman Khan}}

\affiliation[]{Mohamed bin Zayed University of Artificial Intelligence}

\abstract{
Video generators and video-based world models can synthesize plausible visual trajectories, but long-horizon procedural tasks require generation to adapt to what has actually been produced. A model must determine the next action from its generated state, execute that action, and recognize when the task is complete. Open-loop generation cannot adapt to execution outcomes, while existing closed-loop systems often rely on pretrained executors or indirect verification. This leaves a gap between deciding an action and successfully realizing it.
We formulate procedural video generation as \emph{closed-loop task execution in visual world space} and introduce \textbf{WorldGuide}. Given only an initial image and a task goal, WorldGuide predicts an atomic action, generates its corresponding video clip, and uses the generated result to select the next action or terminate. The Planner and Executor are trained on the same step-level procedural demonstrations: the Planner learns to predict the next atomic action or task completion from visual progress, while the Executor is directly trained to realize the predicted actions. Hierarchical visual memory maintains state across long-horizon execution with bounded history token cost.
Due to the lack of step-level action-video supervision for joint planner-executor training, we introduce \textbf{WorldGuide Bench}: approximately 59K step-annotated videos across 245 tasks and 27 procedural categories. WorldGuide achieves a 33.33\% Task Success on \textbf{WorldGuide-Bench}, compared with 29.90\% for the strong recent video model MiniMax-H3, even though MiniMax-H3 receives reference action plans, and achieves 47.69\% on \textbf{VideoCraft-Bench} compared with 32.73\% for MiniMax-H3 under goal-only conditioning. These results demonstrate the importance of coupling planning with learned execution for goal-directed procedural video generation. \\

\vspace{2pt}
\noindent
\faGlobe\ \textbf{Project Page:} \href{https://mbzuai-oryx.github.io/WorldGuide/}{mbzuai-oryx.github.io/WorldGuide}
\\
\faGithub\ \textbf{GitHub:} \href{https://github.com/mbzuai-oryx/WorldGuide}{mbzuai-oryx/WorldGuide}
}

\begin{document}

\maketitle

\section{Introduction}
\label{sec:introduction}

\begin{figure}
    \centering
    \scalebox{1}[1.0]{%
        \includegraphics[width=1.0\linewidth]{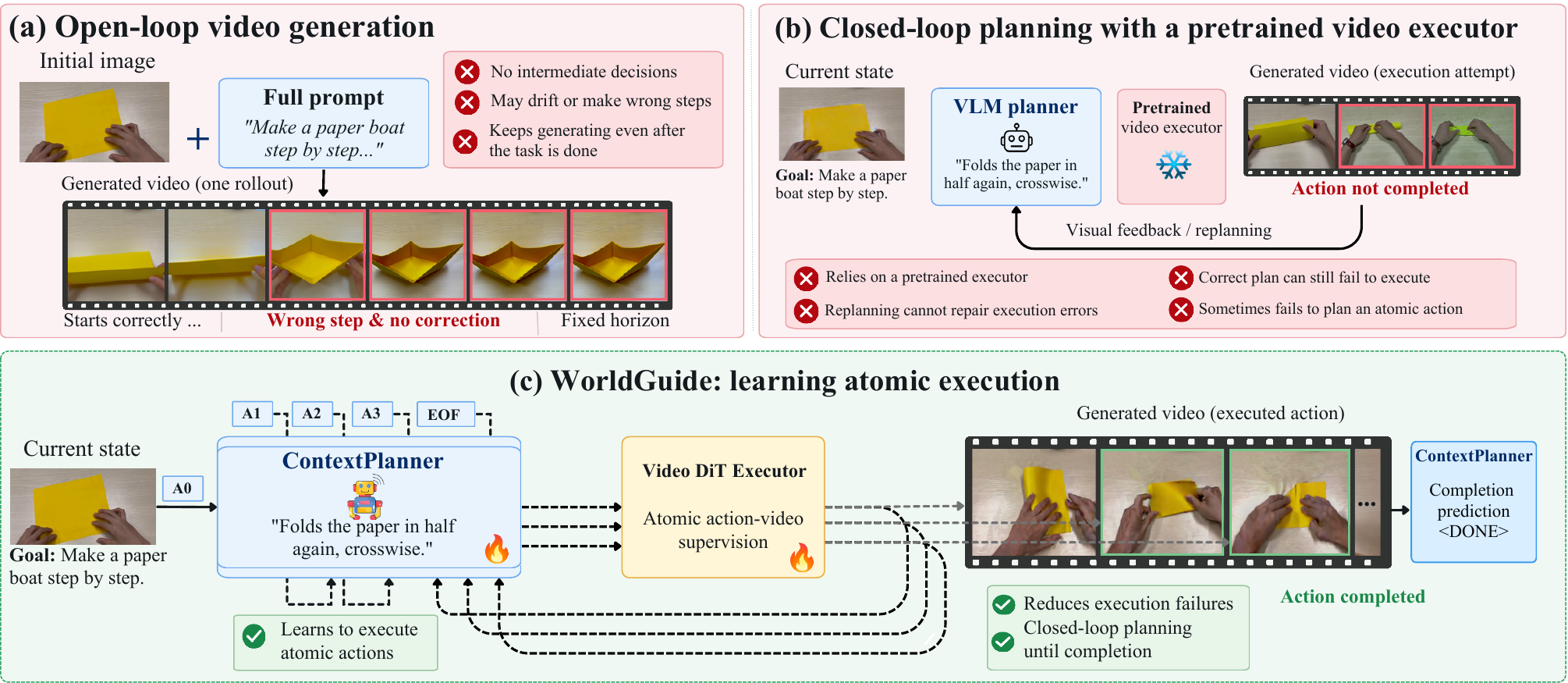}%
    }
    
    \caption{\textbf{From open-loop synthesis to learned procedural execution.} (a) Open-loop generation follows a fixed prompt with no intermediate decisions, so it can drift into a wrong step and keep generating after the task is done. (b) A planner can inspect outcomes and replan, but with a frozen pretrained executor, a correct instruction can still fail to execute. (c) WorldGuide trains the ContextPlanner and Executor on the same demonstrations, the Executor conditioned on the planner's action embeddings. Thus, generated outcomes guide the next action and the decision to stop.}
    \label{fig:HRWM_fig}
\end{figure}

Video-based world models predict how visual states evolve under actions, goals, or other control signals. Direct prompt-to-video models~\citep{openai2025sora2,deepmind2025veo,wu2025hunyuanvideo,kong2024hunyuanvideo,wan2025wan,yang2025cogvideox} and autoregressive approaches~\citep{yuan2025lumos,deng2024nova} can synthesize plausible visual trajectories, but visual plausibility does not ensure task completion. Open-loop generation fixes the prompt or action sequence before rollout, without observing progress, revising subsequent actions, or determining when the goal is reached. As a result, an incorrect or incomplete step cannot be corrected and generation may continue beyond task completion (Fig.~\ref{fig:HRWM_fig}a). This becomes particularly problematic for long-horizon procedures such as origami, assembly, and cooking, where errors in one step can affect everything that follows.

Closed-loop approaches introduce feedback, but a gap remains between planning an action and reliably executing it. ORCA~\citep{he2025orca}, CollabVR~\citep{kim2026collabvr}, and NovaPlan~\citep{fu2026novaplan} pair a pretrained vision-language planner with a frozen video executor. A correct plan can therefore still fail when the executor cannot realize the requested atomic action (Fig.~\ref{fig:HRWM_fig}b). SPIRAL~\citep{yang2026spiral} uses a separately trained critic to evaluate generated outcomes and provides its judgment, rather than directly supervising atomic action execution. Bernini~\citep{team2026bernini} plans in latent semantic space before rendering, but does not predict successive procedural actions from generated progress or determine task completion. Goal-directed generation therefore requires a planner grounded in the evolving visual state and an executor trained to realize the actions it predicts. This motivates our formulation of \emph{closed-loop task execution in visual world space}.

This gap raises five challenges that a closed-loop procedural video model must address. First, if the executor is frozen or only indirectly supervised through a reward model, it cannot be trusted to correctly realize an arbitrary planner-proposed action; \emph{the executor itself needs to be trained on the atomic actions it is asked to perform}. Second, a planner conditioned only on its previously requested actions cannot determine whether those actions were successfully executed, partially completed, or resulted in visual drift. Therefore, \emph{planning must be grounded in the generated visual state, not action history alone}. 
Third, \emph{task completion must be predicted from visual progress rather than determined by a fixed generation length or external retry budget.}
Fourth, because every generated clip affects subsequent decisions, \emph{long-horizon execution requires persistent visual context without unbounded memory growth}. Fifth, existing instructional-video datasets do not provide demonstrations that pair atomic actions with their visual consequences and explicit completion signals. Therefore, \emph{closed-loop procedural training requires dedicated step-level action--video supervision}.

To address these challenges, we introduce \textbf{WorldGuide}, a video world model for closed-loop procedural task execution (Fig.~\ref{fig:HRWM_fig}c). First, we construct \textbf{WorldGuide Bench} (\S~\ref{subsec:dataset_construction}), comprising approximately 59K procedural videos across 245 tasks and 27 categories, segmented into temporally grounded atomic action clips with explicit completion signals. Second, we train the \textbf{ContextPlanner} (\S~\ref{subsec:planner}) on this data to predict the next atomic action or a completion token from the generated visual state and action history (\S~\ref{subsec:overview}). Third, we train the \textbf{Executor} (\S~\ref{subsec:executor}) on the same demonstrations, conditioned on the frozen ContextPlanner's action embeddings, so it learns to render the same atomic actions the planner is trained to propose. Fourth, to support long-horizon state continuity, the Executor maintains a \textbf{hierarchical, compressed visual memory} (\S~\ref{subsec:memory}), representing older history at progressively coarser spatial resolution to bound the token cost as the rollout grows. 
Our experiments validate these design choices. Closed-loop execution improves Task Success by 21.62\% over the open-loop variant (\S~\ref{sec:quantitative_results}); visual feedback improves Task Success by 18.61\% (\S~\ref{sec:visual_feedback}) and Plan Success by 5.34\% while reducing Repeat/Skip by 4.46\% (\S~\ref{sec:planner_ablation}); while memory provides gains of 5.77\% on WorldGuide Bench and 17.91\% on Video-CraftBench (\S~\ref{sec:error_interaction}--\S~\ref{sec:quality_tradeoffs}). Our proposed approach achieves the highest Aesthetic Quality and Overall Consistency on WorldGuide Bench (\S~\ref{sec:qualitative_results}).




\section{Related Work}
\label{sec:related_work}

\subsection{Video Generation and World Models}
\label{subsec:related_video_world}

Modern video generators~\citep{yang2025cogvideox,wan2025wan,wu2025hunyuanvideo, deepmind2025veo} synthesize strong visual trajectories, but operate open-loop, without selecting the next semantic action from generated progress or recognizing task completion. Long-context methods such as FramePack~\citep{zhang2026frame} and interactive world generators~\citep{mao2025yumeoriginal,mao2025yume} improve temporal continuity via compressed visual history, but memory alone does not determine how to advance a procedure. World models extend visual prediction toward decision making: VideoWorld~2~\citep{ren2026videoworld} learns policies through autoregressive latent dynamics, hierarchical world models~\citep{zhang2026hierarchical} plan over latent subgoals, vision-language-action models~\citep{zitkovich2023rt,kim2024openvla} predict robot controls, and Video Language Planning~\citep{du2024videolanguage} combines policies, value functions, and video dynamics through tree search. None of these approaches select actions from visual progress, execute them through directly supervised video generation, and decide completion within a single closed loop. WorldGuide addresses this by coupling visual memory with progress-conditioned action selection, atomic-action execution, and learned termination, all grounded in generated visual states.

\subsection{Planner--Executor Video Generation}
\label{subsec:related_planner_executor}

Planner--executor systems move video generation toward closed-loop interaction, but couple planning and execution differently. BERNINI~\citep{team2026bernini} plans latent visual semantics before rendering, without recurrent procedural decisions from generated progress. TempAct~\citep{wang2026tempact} jointly trains planning and execution but follows prescribed temporal spans without visual-state replanning. PhysAgent~\citep{li2026physagent} refines physical programs via scene reconstruction and physics simulation before synthesis. ORCA~\citep{he2025orca} and CollabVR~\citep{kim2026collabvr} add visual feedback through verification and retries, but keep pretrained executors untrained with their planners. SPIRAL~\citep{yang2026spiral} trains planning and generation jointly, but still relies on a separate critic and reinforcement learning for execution refinement. None learn the full procedural loop: selecting an action from generated progress, executing it under direct supervision, and deciding completion. WorldGuide instead learns next-action selection, atomic-action execution, and completion prediction from the same demonstrations, coupling planning with learned execution in a recurrent closed loop.
\section{Method}
\label{sec:method}

\subsection{Overview and Problem Formulation}
\label{subsec:overview}

\begin{figure}
    \centering
    \includegraphics[width=1.0\linewidth]{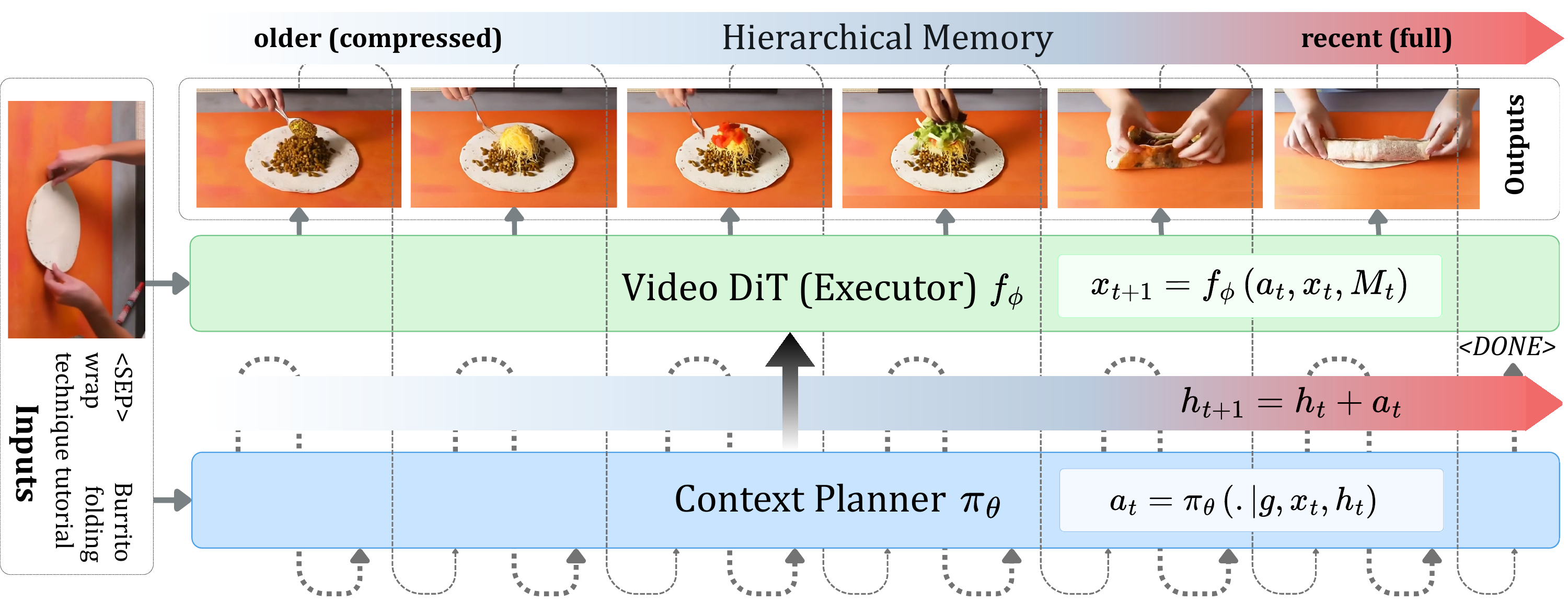}
    \caption{\textbf{Overview of WorldGuide.} Given the initial image $x_0$, task goal $g$, and the latest $K{=}3$ clip--action pairs $h_t$, the ContextPlanner $\pi_\theta$ predicts the next atomic action $a_t$ or $\mathtt{DONE}$. The Executor $f_\phi$ renders $a_t$ into the next clip $x_{t+1}$ from the current visual state $x_t$ and hierarchical memory $M_t$, which progressively compresses older latent frames. Each new clip $x_{t+1}$ becomes the next visual state, is paired with $a_t$ in the history $h_{t+1}$, and updates the memory $M_{t+1}$, and the loop continues until $\mathtt{DONE}$ is predicted or the step budget is reached. Notation follows Eq.~\ref{eq:rollout}.}
    \label{fig:architecture}
\end{figure}

Given an initial image $x_0$ and task goal $g$, \textbf{WorldGuide} performs goal-directed video world modeling through a recurrent planner--executor loop (Fig.~\ref{fig:architecture}). At step $t$, the ContextPlanner $\pi_\theta$ predicts $a_t \in \mathcal{A}\cup\{\mathtt{DONE}\}$ from the goal $g$, the current visual state $x_t$, and the clip--action history $h_t$, where $\mathcal{A}$ is the atomic-action space and $\mathtt{DONE}$ is the completion token \texttt{<|Task Completed|>}. If $a_t=\mathtt{DONE}$, the rollout terminates; otherwise, the Executor $f_\phi$ generates the next visual state $x_{t+1}$, a short video clip, conditioned on $a_t$, $x_t$, and visual memory $M_t$:

\begin{equation}
\label{eq:rollout}
\begin{gathered}
a_t \sim \pi_\theta\!\left(\cdot \mid g,\,x_t,\,h_t\right),
\qquad
x_{t+1}=f_\phi\!\left(a_t,\,x_t,\,M_t\right), \\[4pt]
h_{t+1}=h_t+a_t,
\qquad
M_{t+1}=\mathrm{update}\!\left(M_t,x_{t+1}\right)
\end{gathered}
\end{equation}

with $h_0$ and $M_0$ empty. For $t\geq1$, $x_t$ is the clip generated for $a_{t-1}$, and the Executor uses its last frame as the reference image. The history $h_t$ holds the latest $K{=}3$ actions, each paired with the clip it generated: $h_t+a_t$ appends $a_t$ together with its clip $x_{t+1}$ and drops pairs older than $K$ steps. $\mathrm{update}$ adds $x_{t+1}$ to the Executor's memory. The rollout ends when the planner predicts $\mathtt{DONE}$ or the step budget is reached, and the generated clips $x_1,x_2,\ldots$ are concatenated into the final video (Appendix~\ref{app:inference}).

The ContextPlanner and Executor are initialized from Qwen2.5-VL-7B and HunyuanVideo-1.5 respectively. We train the ContextPlanner first, freeze it, and use it to embed action captions for Executor fine-tuning (Appendix~\ref{app:design_rationale}); at inference, action prediction and termination depend on the Executor's generated results. The Executor also adapts an existing hierarchical memory design (\S~\ref{subsec:memory}) for visual continuity.

\subsection{ContextPlanner}
\label{subsec:planner}

The planner predicts the next atomic action $a_t$ from the task goal $g$, the current visual state $x_t$, and the history $h_t=\big((a_i,x_{i+1})\big)_{i=\max(0,t-K)}^{t-1}$, which pairs each recent action with the clip it generated; for $t>0$, $x_t$ is the newest clip in $h_t$. Thus, $K=1$ supplies one clip and its action, and $K=3$ supplies the last three clips and actions. We use $K=3$. This history is distinct from the Executor's compressed latent memory $M_t$, which the planner does not consume. The output is a single natural-language instruction or a completion token; the visual history provides evidence of recent progress and helps avoid repeated steps.
We train it by supervised fine-tuning of Qwen2.5-VL-7B on demonstration histories and their next-action or completion targets, using token-level cross-entropy masked to response tokens. No architectural modification or auxiliary objective is used.

\subsection{Executor}
\label{subsec:executor}

The Executor is a HunyuanVideo-1.5~\citep{wu2025hunyuanvideo} diffusion transformer conditioned on action embeddings from the frozen ContextPlanner and on visual context. During training, it receives the ground-truth action caption paired with the target clip; at inference, it receives the ContextPlanner's predicted action through the same embedding pathway.

Each training sample provides a target video latent $z \in \mathbb{R}^{C \times L \times H \times W}$, where $L$ is the clip's latent-frame length, with text and vision conditioning features $c_{\text{text}}, c_{\text{vision}}$. The text condition $c_{\text{text}}$ includes the ground-truth action embeddings from the frozen ContextPlanner. Under a flow-matching parameterization, the noisy latent at noise level $\sigma$ is $\tilde{z} = (1-\sigma)z + \sigma \epsilon$ with $\epsilon \sim \mathcal{N}(0, I)$, and the model predicts the residual velocity $\hat{y}_\phi = f_\phi(\tilde z, \sigma, c)$ toward the target $(\epsilon - z)$:
\begin{equation}
\label{eq:exec}
\mathcal{L}_{\text{exec}} =
\mathbb{E}_{z,\epsilon,\sigma}
\left[
\frac{
\left\|
m \odot
\left(
\hat{y}_{\phi} - (\epsilon - z)
\right)
\right\|_2^2
}{
\max\!\left(\sum m, 1\right)
}
\right],
\end{equation}
where $m$ restricts supervision to valid regions. This loss updates only Executor parameters $\phi$; ContextPlanner parameters $\theta$ remain frozen. Reference-image conditioning enters as a latent tensor $z_{\mathrm{cond}} \in \mathbb{R}^{C \times L \times H \times W}$ stacked channel-wise with $\tilde z$, accompanied by a validity mask $m_{\mathrm{cond}}$, so that $c = (z_{\mathrm{cond}}, m_{\mathrm{cond}}, c_{\text{text}}, c_{\text{vision}})$. Without memory, only the reference frame is populated and the remaining slots are masked out. With memory, history latents also populate the conditioning tensor and are embedded into context tokens (Sec.~\ref{subsec:memory}; Appendix~\ref{app:memory_embedders}).

\subsection{Hierarchical Memory Compression}
\label{subsec:memory}

\begin{wrapfigure}{r}{0.60\textwidth}
\centering
\vspace{-1em}
\includegraphics[width=\linewidth,trim={0 25 0 70},clip]{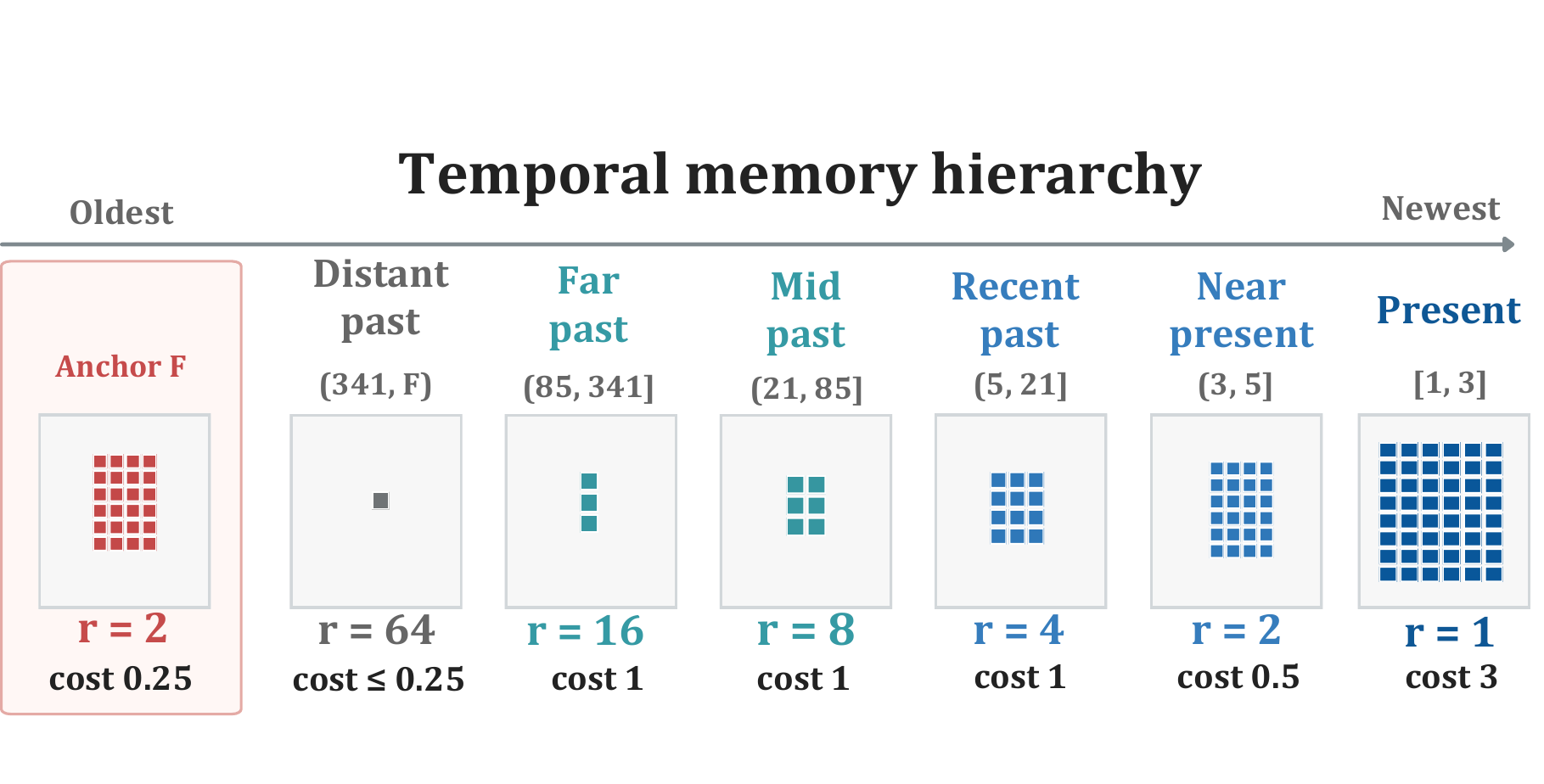}
\caption{Longest-history branch ($343\leq F\leq1366$). Age $1$ is newest; age $F$ is the anchor. Rates $r$ denote spatial reduction relative to base patch embedding, and costs are nominal latent-frame equivalents before spatial padding, with the separately embedded reference adding cost $1$ at $r=1$. Shorter branches appear in Appendix~\ref{app:memory_budget}.}
\label{fig:memory_tiers}
\vspace{-1em}
\end{wrapfigure}

We adopt YUME's spatial history-compression schedule, following Yume-1.5~\cite{mao2025yume} and FramePack~\citep{zhang2026frame}. Recent latent frames receive finer spatial embeddings than older ones, bounding the Executor's history-token count.

\noindent\textbf{Memory bank.} Let $T\leq1365$ denote the capacity for non-anchor history latent frames. The compressor additionally handles the oldest history latent frame as an anchor, giving a total history length $F\leq T+1\leq1366$:
\begin{equation}
\label{eq:membank}
\scalebox{0.9}{$\displaystyle
M_t = \Pi_T\!\left(\left[\, z^{(1)} \,\|\, \cdots \,\|\, z^{(t)} \,\right]\right)
$}
\end{equation}

Here, $z^{(i)}\in \mathbb{R}^{C \times F \times H \times W}$ is the VAE latent of the generated clip $x_i$, $\Pi_T$ forms the bounded history tensor from these preceding clips, and $H,W$ are latent spatial dimensions. $F$ counts history latent slots supplied to compression, including the anchor and temporal padding; the current reference is separate. These are latent-frame counts after VAE encoding, not video frames. The schedule capacity is an upper limit, not the length used at every step; Appendix~\ref{app:memory_bank} specifies how the conditioning path limits $F$.

\noindent\textbf{Multi-scale patch embedding.} With DiT spatial patch size $p=2$, $E_r$ is a 3-D convolution with kernel and stride $(1,pr,pr)$ for $r\in\{1,2,4,8,16\}$. It maps $C$ latent channels to $d$ features. We initialize $E_1$ from the pretrained patch embedding and larger kernels by trilinear upsampling. The coarsest embedder is $E_{64}=E_{16}\circ\Phi$, where $\Phi$ is a channel-preserving convolution with kernel and stride $(1,4,4)$. Its effective spatial stride is $128=64p$. All embedders preserve temporal length.


\noindent\textbf{Partition and token budget.} For nonempty history, the selected branch splits it into segments $S_j$ with spatial rates $r_j$, the anchor at rate $r_a$; Fig.~\ref{fig:memory_tiers} shows the longest branch. Ignoring spatial padding,
\begin{equation}
\label{eq:segment}
\begin{aligned}
h_j &= \mathrm{flatten}\!\left(E_{r_j}\!\left(M_t[:,S_j]\right)\right)
\in\mathbb{R}^{n_j\times d}, 
n_j &= |S_j|\frac{HW}{p^2r_j^2}.
\end{aligned}
\end{equation}
The anchor, chronological history segments, and current reference form

\begin{equation}
\mathcal{M} = [\,h_a\,;\,h_1\,;\,\cdots\,;\,h_J\,;\,\mathrm{flatten}(E_1(z_{\mathrm{ref}}))\,],
\end{equation}
where $h_a$ embeds the oldest history latent frame. Across all supported branches,
\begin{equation}
\label{eq:budget}
\frac{p^2}{HW}|\mathcal{M}| = \frac{1}{r_a^2}+\sum_j\frac{|S_j|}{r_j^2}+1\leq8<9,
\qquad T\leq1365.
\end{equation}

\begin{wrapfigure}{r}{0.48\textwidth}
\vspace{-1em}
    \centering
    \includegraphics[width=\linewidth]{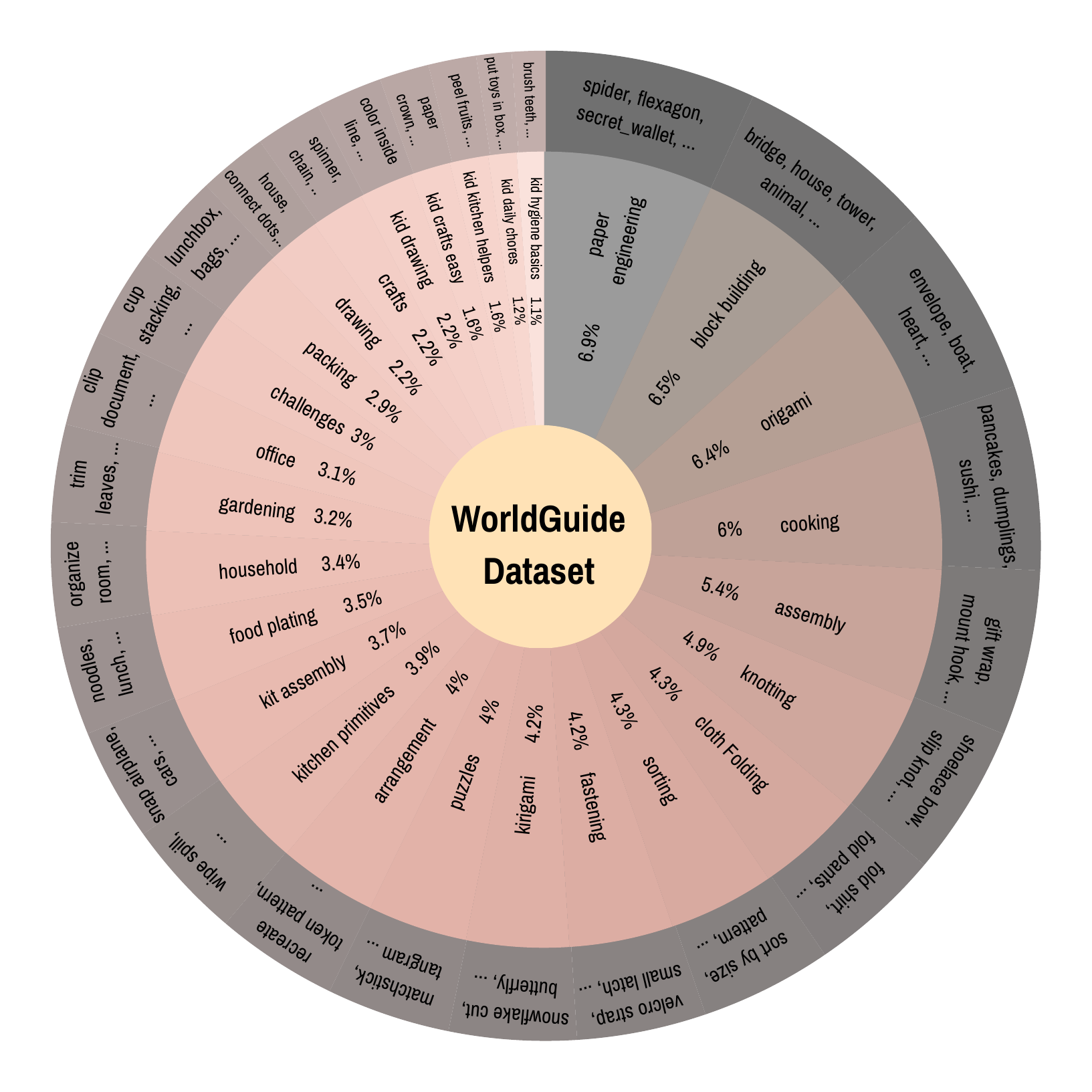}
    \caption{Task taxonomy of WorldGuide Bench. The inner ring shows procedural categories and their dataset shares; the outer ring lists representative tasks within each category.}
    \label{fig:dataset_distribution}
\vspace{-1em}
\end{wrapfigure}


One full-resolution latent-frame equivalent is $HW/p^2$ tokens. This bound measures token cost, including the anchor and reference, not a reduction of history to eight temporal latent frames; spatial padding changes the actual count (Appendix~\ref{app:memory_budget}).




\subsection{Dataset for WorldGuide Training}
\label{subsec:dataset_construction}

\begin{table*}[t]
\centering
\caption{
\textbf{Procedural video datasets and their supervision for planning and generation.}
Sizes retain their original units: videos (vid.), segments (seg.), clips.
Tasks/Dom.: tasks/domains;
Step-Clip: temporally localized step annotations;
Atomic Act.: single-action granularity;
Coverage Target: annotation protocol targets all visible task-relevant steps, excluding background intervals, rather than a verified coverage rate;
Task Goal: explicit task-level goal;
Completion: explicit task-completion supervision.
Gen., Eval., and Recog.\ denote generation, evaluation, and recognition.
}
\label{tab:dataset_comparison}

\resizebox{1.0\textwidth}{!}{%
\begin{tabular}{lccccccccc}
\toprule
\textbf{Dataset} & \textbf{Size} & \textbf{Tasks/Dom.} & \textbf{Step-Clip} & \textbf{Atomic Act.} & \textbf{Coverage Target} & \textbf{Task Goal} & \textbf{Completion} & \textbf{Type} & \textbf{Public} \\
\midrule

EgoPlan-IT & 50K QA & --/1 & \xmark & \cmark & \xmark & \cmark & \xmark & Planning QA & \cmark \\
Ego4D Goal-Step & 48K seg. & 86/1 & \cmark & \xmark & \xmark & \cmark & \xmark & Localization & \cmark \\
COIN & 11.8K vid. & 180/12 & Fixed-vocab & Partial & \xmark & \cmark & \xmark & Localization & \cmark \\
CrossTask (primary) & 2.75K vid. & 18/4 & \cmark & Partial & \xmark & \cmark & \xmark & Weak localization & \cmark \\
YouCook2 & 2K vid. & 89/1 & \cmark & \xmark & \cmark & \cmark & \xmark & Segmentation & \cmark \\
HT-Step & 19.7K vid. & 433/1 & \cmark & Partial & \xmark & \cmark & \xmark & Grounding & \cmark \\
Assembly101 & 4.3K vid. & 101/1 & \cmark & \cmark & \cmark & \xmark & \xmark & Recognition & \cmark \\
CaptainCook4D & 384 vid. & 24/1 & \cmark & Partial & \cmark & \cmark & \xmark & Error/localization & \cmark \\
SemComp-Data & 1.27K & 21/6 & \xmark & \xmark & \xmark & \cmark & \xmark & Eval-only & Partial \\
ActVideoGen & 118K clips & --/Multi & \cmark & Partial & Partial & \cmark & \xmark & Gen.\ training & \xmark \\
EgoForge / X-Ego & 15K clips & --/Multi & \xmark & \cmark & \xmark & \cmark & \xmark & Gen. training + eval & \xmark \\
Ego-Exo4D Keystep & 27.6K seg. & 17/3 & \cmark & Partial & \xmark & \cmark & \xmark & Recognition & \cmark \\

\midrule

\textbf{WorldGuide (ours)} & \textbf{59K vid.} & \textbf{245/27} & \textbf{\cmark} & \textbf{\cmark} & \textbf{\cmark} & \textbf{\cmark} & \textbf{\cmark} & \textbf{Gen.\ training + eval} & \textbf{Planned} \\

\bottomrule
\end{tabular}%
}
\end{table*}

Closed-loop procedural execution requires supervision for three connected decisions: which action follows observed progress, how that action changes the visual state, and when the task is complete. Existing instructional datasets provide narration, action labels, or localized step descriptions~\citep{miech2019howto100m,zhou2018youcook2,tang2019coin,zhukov2019crosstask,damen2018epickitchens}, but need additional processing to support this training formulation. We construct \textbf{WorldGuide Bench} to combine ordered atomic-action clips, task goals, and explicit completion signals within each demonstration (Table~\ref{tab:dataset_comparison}). This shared supervision supports both ContextPlanner decisions and Executor training.

The dataset contains \textbf{58,679 videos} spanning \textbf{245 tasks} across \textbf{27 procedural categories} (Fig.~\ref{fig:dataset_distribution}). We use Gemini~2.5~Flash~\citep{gemini25} to filter web instructional videos and annotate temporally grounded atomic actions and completion. We used total \textbf{980 videos} in test dataset. Construction details, annotation examples, and training-sample preparation are provided in Appendix~\ref{app:dataset_details}, and a human audit of the annotations in Appendix~\ref{app:human_dataset_audit}.


\section{Experiments}
\label{sec:experiments}

Open-loop video generation follows instructions and a duration chosen before rollout, without checking intermediate task progress or deciding when the goal has been reached. Planner--executor systems can introduce feedback, but some use pretrained components without procedural training, while others plan the action sequence before rendering. WorldGuide trains both the ContextPlanner and Executor on procedural demonstrations to select and carry out atomic actions toward a long-horizon goal. Generated outcomes then guide the next action and the decision to stop.

This formulation raises five empirical questions. \textit{First}, does closed-loop execution improve procedural task completion? \textit{Second}, does generated visual feedback improve replanning? \textit{Third}, can WorldGuide recognize completion and stop? \textit{Fourth}, how do planning and execution failures interact? \textit{Fifth}, does improving procedural correctness come at the cost of visual quality?

\subsection{Experimental Setup}
\label{sec:experimental_setup}

We fine-tune Qwen2.5-VL-7B-Instruct as the ContextPlanner and HunyuanVideo-1.5 as the Executor, on WorldGuide Bench. The planner uses $K{=}3$ recent clip-action pairs. We evaluate 980 test samples from WorldGuide Bench and all 294 Video-CraftBench samples~\citep{ren2026videoworld}. Training and inference settings are given in Appendix~\ref{app:training_details}.
On WorldGuide Bench, starred baselines (Bernini, PhysAgent, and TempAct) use their own planning-execution loops. WorldGuide receives only the initial image and task goal. On Video-CraftBench, all models receive the initial image and goal without reference actions. Appendix~\ref{app:eval_protocol} details the protocols and ablations.

\begin{table}[!t]
\centering

\caption{\textbf{WorldGuide Bench evaluation.} Unstarred baselines use reference actions; $*$ denotes methods with their own planning-execution loop. WorldGuide predicts actions from the task goal and generated history. \textbf{Aesth.}: Aesthetic Quality; \textbf{Imag.}: Imaging Quality; \textbf{Dynamic}: Dynamic Degree; \textbf{Smooth.}: Motion Smoothness; \textbf{Consist.}: Overall Consistency; \textbf{Align.}: Alignment; \textbf{Plan Acc.}: Plan Accuracy; \textbf{Task Succ.}: Task Success Rate. Higher is better except Repeat/Skip (Appendix~\ref{app:eval_metrics}).}

\label{tab:main_results}
\resizebox{1.0\linewidth}{!}{
\begin{tabular}{l|ccccc|ccccc}
\toprule
\multirow{2}{*}{Model} 
& \multicolumn{5}{c|}{Video Quality Check} 
& \multicolumn{5}{c}{Video Planning Check} \\
\cmidrule(lr){2-6} \cmidrule(lr){7-11}
& Aesth.
& Imag.
& Dynamic
& Smooth.
& Consist.
& Align.
& Plan Acc.
& Order
& Repeat/Skip $\downarrow$
& Task Succ. \\

\midrule
Cosmos-Predict2.5-2B 
& 38.77 & 62.11 & 82.52 & 96.32 & 17.46 
& 63.75 & 38.05 & 84.08 & 3.98 & 11.94 \\

Open-Sora 2.0-11B
& 33.42 & 44.91 & 60.19 & 95.06 & 17.42
& 62.43 & 31.27 & 79.56 & 6.53 & 8.54 \\

HunyuanVideo-1.5-8.3B
& 42.43 & 65.57 & 88.84 & 96.00 & 18.91
& 62.28 & 43.53 & 80.00 & 3.00 & 12.00 \\

Wan2.2-14B
& 41.46 & 58.87 & 16.02 & 93.67 & 17.52
& 67.33 & 4.37 & 21.65 & 2.06 & 2.58 \\

CogVideoX-5B
& 38.76 & 54.99 & 64.08 & 96.58 & 18.09
& 68.30 & 34.33 & 75.74 & 3.96 & 9.90 \\

Helios-14B
& 39.32 & 56.06 & 72.33 & 95.90 & 18.60
& 63.99 & 25.71 & 76.47 & 1.51 & 8.04 \\

UniVideo-13B
& \underline{45.75} & 61.92 & 48.00 & 97.09 & 18.05
& 65.63 & 10.77 & 52.00 & 4.00 & 4.00 \\

FlashMotion-34B
& 41.69 & \underline{67.26} & 90.29 & 95.46 & 17.93
& 62.92 & 24.10 & 77.39 & 2.51 & 5.53 \\

MAGI-1-24B
& 39.14 & 63.08 & 2.30 & 98.31 & 17.00
& 69.33 & 17.70 & 61.63 & 2.33 & 3.49 \\

RoboMaster-5B
& 32.60 & 17.68 & 4.37 & 97.09 & 6.66
& 37.13 & 4.29 & 5.50 & \underline{0.50} & 4.00 \\

SpMem-5B
& 35.79 & 47.69 & 96.73 & 98.94 & \underline{20.72}
& 60.92 & 33.74 & 78.95 & 2.94 & 10.53 \\

Astra-1.3B
& 32.00 & 43.43 & 97.67 & 95.73 & 15.03
& 65.40 & 27.23 & 60.71 & 0.64 & 11.90 \\

Yume-1.5-5B
& 40.61 & 65.72 & 87.86 & 96.90 & 17.39
& 60.79 & 19.72 & 74.26 & 2.39 & 2.97 \\

SkyReels-V3-14B
& 42.64 & 58.78 & 14.56 & \textbf{99.66} & 17.03
& \underline{72.54} & 11.23 & 48.52 & 1.01 & 3.90 \\

MiniMax-H3-33B\footnotemark
\edef\minimaxfootnote{\thefootnote}
& 41.61 & 56.60 & 80.58 & 97.99 & 19.84
& 70.04 & \underline{57.99} & \underline{86.79} & 4.06 & \underline{29.90} \\

LTX-2.5-22B
& 43.85 & 57.92 & 85.92 & \underline{99.29} & 15.63
& 70.10 & 16.14 & 49.76 & 1.46 & 7.32 \\

HY-WorldPlay-8.3B
& 39.25 & 64.56 & 85.95 & 96.30 & 15.28
& 55.47 & 15.67 & 73.46 & 3.07 & 1.26 \\

PhysAgent*
& 42.76 & 65.09 & 52.68 & 98.39 & 16.72
& 70.61 & 18.20 & 56.56 & 1.98 & 8.42 \\

TempAct-1.3B*
& 44.68 & \textbf{71.06} & \textbf{100.00} & 97.81 & 15.23
& 65.09 & 9.11 & 29.68 & \underline{0.50} & 3.98 \\

Bernini-14B*
& 45.12 & 63.13 & 93.20 & 98.23 & 11.91
& 53.48 & 5.30 & 22.06 & \textbf{0.00} & 1.96 \\

\midrule

\rowcolor{lightblue}
WorldGuide-8.3B (ours)
& \textbf{49.36}
& 64.77
& \underline{98.06}
& 98.96
& \textbf{31.13}
& \textbf{73.85}
& \textbf{62.85}
& \textbf{92.42}
& 6.06
& \textbf{33.33} \\
\bottomrule
\end{tabular}
}
\end{table}

\footnotetext{\url{https://huggingface.co/MiniMaxAI/MiniMax-H3}}

\begin{table}[ht]
\centering
\caption{\textbf{Video-CraftBench: goal-conditioned generation.} All models receive the initial image and task goal without reference actions. $*$ denotes methods with their own planning--execution loops. Higher is better except Repeat/Skip.}
\label{tab:main_results2}
\resizebox{1.0\linewidth}{!}{
\begin{tabular}{l|ccccc|ccccc}
\toprule
\multirow{2}{*}{Model} 
& \multicolumn{5}{c|}{Video Quality Check} 
& \multicolumn{5}{c}{Video Planning Check} \\
\cmidrule(lr){2-6} \cmidrule(lr){7-11}
& Aesth.
& Imag.
& Dynamic
& Smooth.
& Consist.
& Align.
& Plan Acc.
& Order
& Repeat/Skip $\downarrow$
& Task Succ. \\
\midrule

Cosmos-Predict2.5-2B 
& 35.83 & 63.66 & 95.92 & 98.61 & 19.49 
& 63.54 & 5.66 & 32.42 & 5.46 & 2.05 \\

Open-Sora 2.0        
& 33.86 & 30.26 & 47.62 & 99.04 & 21.61 
& 52.97 & 22.03 & 82.96 & 2.05 & 14.32 \\

HunyuanVideo-1.5     
& \underline{45.30} & 67.62 & \underline{99.66} & 98.91 & 21.91 
& 63.94 & 14.66 & 64.97 & 3.06 & 24.15 \\

Wan2.2               
& \textbf{46.05} & \underline{72.47} & \textbf{100.00} & 98.31 & 22.28 
& \underline{71.98} & 2.98 & 18.22 & 4.11 & 2.40 \\

CogVideoX            
& 40.60 & 63.49 & 89.46 & 98.61 & 21.52 
& \textbf{75.13} & 15.04 & 69.73 & 8.16 & 2.38 \\

Helios               
& 36.75 & 49.98 & 92.18 & 99.20 & 19.13 
& 54.76 & 13.83 & 80.56 & \underline{1.36} & 6.46 \\

UniVideo             
& 45.05 & 51.71 & 62.46 & 97.52 & 20.99 
& 60.41 & 9.77 & 50.55 & 4.03 & 12.09 \\

FlashMotion          
& 43.78 & 67.29 & 99.32 & 98.83 & 20.73 
& 59.19 & 7.32 & 38.89 & 1.70 & 3.74 \\

MAGI-1               
& 36.72 & \textbf{75.40} & 0.00 & \textbf{99.61} & 16.40 
& 71.67 & 6.62 & 75.00 & 8.50 & 0.68 \\

RoboMaster           
& 35.98 & 71.30 & \textbf{100.00} & 99.26 & 18.59 
& 65.75 & 3.36 & 15.58 & 1.37 & 0.68 \\

SpMem                
& 36.17 & 62.39 & 94.90 & 99.18 & 20.11 
& 65.49 & 11.72 & 55.11 & 6.46 & 3.74 \\

Astra                
& 27.80 & 38.23 & 95.58 & 94.29 & 7.94 
& 55.56 & 3.11 & 47.96 & 6.80 & 0.00 \\

Yume-1.5             
& 39.99 & 62.40 & \underline{99.66} & 99.26 & 23.13 
& 67.17 & 5.36 & 29.58 & 3.46 & 5.88 \\

SkyReels-V3         
& 42.01 & 72.14 & 41.84 & \underline{99.59} & 19.74 
& 64.50 & 8.29 & 58.80 & 1.74 & 1.04 \\

MiniMax-H3\footnotemark[1]    
& 40.65 & 69.65 & \textbf{100.00} & 99.14 & \underline{23.18} 
& 68.84 & \underline{28.64} & \underline{88.55} & 2.16 & \underline{32.73} \\

LTX-2.5         
& 41.60 & 68.30 & 85.71 & 99.46 & 21.37 
& 66.22 & 6.59 & 46.08 & 11.68 & 3.44 \\

HY-WorldPlay         
& 35.03 & 63.84 & 62.24 & 99.03 & 13.66 
& 66.17 & 2.77 & 38.74 & 1.37 & 0.00 \\

PhysAgent*         
& 39.08 & 67.69 & 88.10 & 97.69 & 18.86 
& 68.61 & 16.62 & 83.13 & 1.40 & 1.05 \\

TempAct*         
& 42.43 & 72.22 & \textbf{100.00} & 98.32 & \textbf{23.54} 
& 64.71 & 7.57 & 51.26 & \textbf{0.34} & 24.47 \\

Bernini*        
& 41.14 & 61.84 & 98.26 & 98.15 & 16.47 
& 58.36 & 5.56 & 29.51 & 1.77 & 7.77 \\

\midrule

\rowcolor{lightblue}
WorldGuide (ours)
& 40.19 & 64.63 & 99.65 & 98.86 & 22.09
& 71.80 & \textbf{57.42} & \textbf{94.45} & 17.08 & \textbf{47.69} \\

\bottomrule
\end{tabular}
}
\end{table}

\paragraph{Metrics.}
\label{sec:metrics}

We assess visual quality and video-text consistency using VBench~\citep{zheng2025vbench}, and measure similarity between generated and reference videos using \textbf{Paired Alignment}. We use Gemini to assess task execution in generated videos. \textbf{Task Success Rate} measures final goal achievement, and \textbf{Plan Accuracy} measures completion of required sub-goals. \textbf{Order Score} evaluates prerequisite order, while \textbf{Repeat/Skip} penalizes unnecessary repetition. Finally, GPT-5.2 evaluates the predicted action sequences. Its \textbf{Plan Success} score indicates whether a sequence would achieve the goal if executed correctly. Full metric definitions and evaluation prompts are provided in Appendix~\ref{app:eval_metrics}.

\subsection{Does Closed-Loop Execution Improve Procedural Task Completion?}
\label{sec:quantitative_results}

Open-loop generation commits to an action sequence and rollout length before execution, preventing subsequent decisions from responding to incomplete or incorrect actions. WorldGuide instead predicts each action-consequences and termination from generated progress. Compared with its open-loop variant, this formulation raises Task Success from 11.71\% to 33.33\% and Plan Accuracy from 22.87 to 62.85 (Table~\ref{tab:ablation_worldguide}). WorldGuide also outperforms MiniMax-H3 on WorldGuide Bench despite its access to reference actions (29.90\% Task Success; Table~\ref{tab:main_results}), and on Video-CraftBench under goal-only conditioning (47.69\% versus 32.73\%; Table~\ref{tab:main_results2}). A human study confirms this ranking (Appendix~\ref{app:human_model_study}).

However, closing the loop alone does not guarantee task completion: the planner must select appropriate atomic actions, and the executor must realize their consequences. As shown in Table~\ref{tab:main_results} and~\ref{tab:main_results2}, WorldGuide consistently outperforms other closed-loop planner--executor methods, including PhysAgent, TempAct, and Bernini, demonstrating that procedural training and grounding each decision in generated outcomes are critical for reliable completion.

\subsection{Does Generated Visual Feedback Improve Replanning?}
\label{sec:visual_feedback}


A closed-loop planner is only useful if it observes what was actually executed. To isolate this effect, we remove generated clips from the planner input, keeping only the task goal and previous actions; the Executor is unchanged. Restoring visual feedback raises Task Success from \textbf{14.72\%} to \textbf{33.33\%} and Plan Accuracy from \textbf{29.29} to \textbf{62.85} (Table~\ref{tab:ablation_worldguide}). This shows action history alone is insufficient: knowing what the planner \emph{requested} does not reveal whether the action was completed, partially realized, or drifted visually. Observing the generated outcome lets WorldGuide replan from the state that exists, so later decisions can compensate for execution errors and stay aligned with task progress.

\begin{table}[t]
\centering
\caption{\textbf{WorldGuide ablations.} Removing step-wise replanning (\textit{open-loop}), planner visual feedback (\textit{w/o visual}), or Executor memory (\textit{w/o Mem}) lowers Task Success; \textit{oracle ContextPlanner} substitutes reference actions to isolate planning errors. Protocols follow Tables~\ref{tab:main_results} and~\ref{tab:main_results2} (Appendix~\ref{app:eval_protocol}).}
\label{tab:ablation_worldguide}
\resizebox{1.0\linewidth}{!}{
\begin{tabular}{l|ccccc|ccccc}
\toprule
\multirow{2}{*}{Variant}
& \multicolumn{5}{c|}{Video Quality Check}
& \multicolumn{5}{c}{Video Planning Check} \\
\cmidrule(lr){2-6} \cmidrule(lr){7-11}
& Aesth.
& Imag.
& Dynamic
& Smooth.
& Consist.
& Align.
& Plan Acc.
& Order
& Repeat/Skip $\downarrow$
& Task Succ. \\
\midrule
\multicolumn{11}{l}{\textit{WorldGuide Bench}} \\

WorldGuide (open-loop)
& 36.76 & 59.38 & 73.30 & 98.42 & 18.86
& 67.40 & 22.87 & 53.81 & 0.49 & 11.71 \\

WorldGuide (w/o visual)
& 36.77 & 58.74 & 83.92 & 98.25 & 18.67
& 68.11 & 29.29 & 68.70 & \textbf{0.08} & 14.72 \\

WorldGuide (w/o Mem)
& 38.70 & 55.35 & 88.35 & 98.14 & 19.70
& 64.74 & 55.36 & 91.12 & 4.39 & 27.56 \\

WorldGuide (oracle ContextPlanner)
& \textbf{52.44}
& \textbf{66.72}
& \underline{97.68}
& \textbf{99.49}
& \textbf{33.37}
& \textbf{77.14}
& \textbf{65.42}
& \textbf{94.28}
& \underline{2.04}
& \textbf{38.82} \\

\rowcolor{lightblue}
WorldGuide (ours)
& \underline{49.36}
& \underline{64.77}
& \textbf{98.06}
& \underline{98.96}
& \underline{31.13}
& \underline{73.85}
& \underline{62.85}
& \underline{92.42}
& 6.06
& \underline{33.33} \\

\midrule
\multicolumn{11}{l}{\textit{Video-CraftBench}} \\

WorldGuide (w/o Mem)
& \underline{38.82}
& \underline{63.81}
& \underline{99.63}
& \underline{98.64}
& \underline{20.21}
& \textbf{73.77}
& \underline{39.62}
& \underline{92.51}
& \textbf{4.10}
& \underline{29.78} \\

\rowcolor{lightblue}
WorldGuide (ours)
& \textbf{40.19}
& \textbf{64.63}
& \textbf{99.65}
& \textbf{98.86}
& \textbf{22.09}
& \underline{71.80}
& \textbf{57.42}
& \textbf{94.45}
& \underline{17.08}
& \textbf{47.69} \\

\bottomrule
\end{tabular}
}
\end{table}

\subsection{Can WorldGuide Recognize Completion and Stop?}
\label{sec:planner_ablation}

Completing a task requires knowing not only what to do next, but when to stop. The same ContextPlanner handles both, predicting either the next atomic action or a completion token from the rollout history. In autonomous execution, using generated visual feedback improves Plan Success from 86.41\% to 91.75\% and reduces Repeat/Skip from 20.19 to 15.73 (Table~\ref{tab:ablation_planner}). This shows that visual observations help the planner distinguish genuine task progress from merely knowing which actions were previously requested, leading to more complete and less redundant procedures.

\begin{table}[t]
\centering
\caption{\textbf{ContextPlanner evaluation on WorldGuide Bench.} Comparison of ground-truth-history planning against autonomous rollout with generated history; visual feedback improves plan quality during autonomous execution. FT: supervised fine-tuning; $K$: history length.}

\label{tab:ablation_planner}
\resizebox{0.85\linewidth}{!}{%
\begin{tabular}{lcccccc}
\toprule
\textbf{Setting} 
& \textbf{FT} 
& \textbf{$K$} 
& \textbf{Plan Acc. $\uparrow$} 
& \textbf{Order $\uparrow$} 
& \textbf{Repeat/Skip $\downarrow$} 
& \textbf{Success $\uparrow$} \\
\midrule

Qwen2.5-VL baseline 
& {\color{cyan}\small\faSnowflake} & 3 & 1.04 & 0.00 & 8.82 & 0.00 \\

ContextPlanner 
& {\color{orange}\small \faFire} & 1 & 56.48 & 42.35 & 21.16 & 39.17 \\

ContextPlanner 
& {\color{orange}\small \faFire} & 2 & 75.83 & 71.28 & 15.56 & 44.66 \\

\rowcolor{lightblue}
ContextPlanner 
& {\color{orange}\small \faFire} & 3 & 79.66 & 72.56 & 14.71 & 47.04 \\

\midrule

WorldGuide (w/o visual) 
& {\color{orange}\small \faFire} & 3 & 63.84 & 77.46 & 20.19 & 86.41 \\

\rowcolor{lightblue}
WorldGuide planner-executor 
& {\color{orange}\small \faFire} & 3 & 64.03 & 79.45 & 15.73 & 91.75 \\

\bottomrule
\end{tabular}%
}
\end{table}

With ground-truth histories, the planner reaches \textbf{79.66} Plan Accuracy at $K=3$, compared with \textbf{64.03} in autonomous rollout, where every decision depends on its generated outcomes. Despite this harder setting, Plan Success still reaches \textbf{91.75\%}, showing WorldGuide can construct and terminate valid procedures from generated progress. Plan Success measures whether the predicted plan would achieve the goal if executed correctly. 
Visual completion is measured separately by the video-judged Task Success (Tables~\ref{tab:main_results}--\ref{tab:ablation_worldguide}).

\begin{figure}
    \centering
    \includegraphics[width=1.0\linewidth]{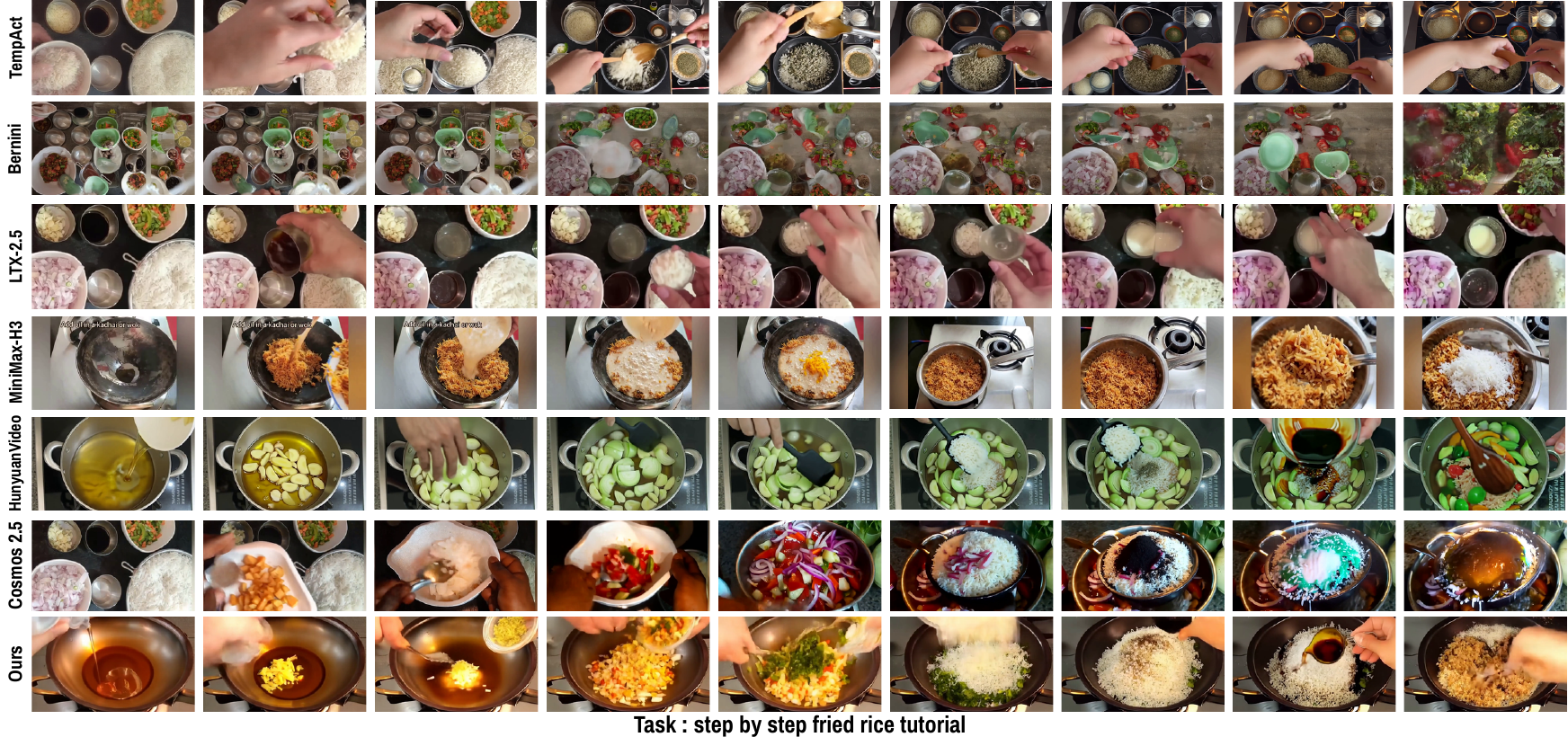}
    \caption{ Qualitative comparison for \textit{``step-by-step fried rice tutorial''}. MiniMax-H3, LTX-2.5, HunyuanVideo, and Cosmos-Predict2.5 receive reference action sequences; Bernini and TempAct use their own planning--execution loops. WorldGuide predicts actions from the task goal and generated progress. Frames are uniformly sampled from each trajectory. }
    \label{fig:qualitative_results}
    \vspace{-1em}
\end{figure}

\subsection{How Do Planning and Execution Failures Interact?}
\label{sec:error_interaction}

Planning and execution errors are tightly coupled in closed-loop generation. Replacing the learned planner with oracle reference actions improves Task Success by \textbf{+5.49 \%} (Table~\ref{tab:ablation_worldguide}), using the same trained Executor and memory, showing stronger planning can push task completion further.


Execution quality also affects what the planner sees next. With the ContextPlanner and Executor checkpoints unchanged, enabling memory improves Task Success by \textbf{+5.77\%} on WorldGuide Bench and \textbf{+17.91\%} on Video-CraftBench, and video-judged Plan Accuracy by \textbf{+7.49\%} (Table~\ref{tab:ablation_worldguide}). Since each generated state feeds the next planning step, execution errors can corrupt later evidence, underscoring how tightly planning and execution depend on each other in closed-loop task completion.

\subsection{Does Procedural Improvement Compromise Visual Quality?}
\label{sec:quality_tradeoffs}

WorldGuide primarily targets reliable procedural completion, so its gains do not uniformly extend to every low-level video-quality metric. On WorldGuide Bench, it achieves the highest Aesthetic Quality (\textbf{49.36}) and Overall Consistency (\textbf{31.13}), while baselines like TempAct and SkyReels score higher on individual metrics such as Imaging Quality, Dynamic Degree, or Motion Smoothness (Table~\ref{tab:main_results}), consistent with differences in backbone scale, pretraining data, and optimization for visual fidelity.

More importantly, WorldGuide substantially improves Task Success, reaching \textbf{33.33\%} on WorldGuide Bench and \textbf{47.69\%} on Video-CraftBench. On Video-CraftBench, this comes with modest drops in some visual-quality metrics relative to MiniMax-H3, including Imaging Quality (\textbf{64.63} vs.\ \textbf{69.65}) and Overall Consistency (\textbf{22.09} vs.\ \textbf{23.18}). WorldGuide thus prioritizes task execution over visual fidelity in challenging cases, and improving this balance remains an important direction.

\subsection{Qualitative Results}
\label{sec:qualitative_results}
In Fig.~\ref{fig:qualitative_results}, WorldGuide adds aromatics, vegetables, and rice before seasoning and mixing. Its text actions identify which ingredients to add and in what order. MiniMax-H3 shows seasoned rice early, then adds white rice again and changes cookware. TempAct and LTX-2.5 largely remain at ingredient handling, while Bernini loses scene consistency. These examples show why plausible individual actions are insufficient for a coherent procedure. Additional tasks appear in Appendix~\ref{app:additional_qualitative}.

\section{Discussion \& Limitations}
\label{sec:discussion}

WorldGuide shows that closed-loop procedural video generation substantially improves task completion, though the remaining gap points to room for more robust long-horizon execution. Failures typically arise from a premature or redundant planner action, or an Executor clip that does not fully realize the intended state transition; since each generated state feeds the next decision, such errors can propagate. The gains from visual feedback and memory show that better state tracking directly improves both planning and execution.

A promising direction is to strengthen this coupling further with richer state-validity signals and explicit recovery when a step is incomplete or inconsistent. The Executor already conditions on the ContextPlanner's action embeddings; joint optimization of both modules is a natural next step. Bounded visual memory compresses older states to control inference cost, and more selective retention of task-relevant information could improve long-horizon consistency while keeping this efficiency. These directions extend the current closed-loop formulation rather than requiring a different one.



\section{Conclusion}
\label{sec:conclusion}

We introduced \textbf{WorldGuide}, a video world model that turns procedural video generation into closed-loop task execution. The central idea is that a procedure should unfold through decisions grounded in the generated visual state: what to do next and when to stop depend on what has actually been accomplished. WorldGuide implements this principle by repeatedly selecting a semantic action, rendering its consequence, and using the resulting visual state to guide subsequent execution. Compressed visual memory supports state continuity across successive steps of long-horizon, goal-directed execution. We also introduced \textbf{WorldGuide Bench}, comprising approximately 59K annotated videos across 245 tasks and 27 procedural categories. Among the evaluated video generators and video world models, WorldGuide achieves the highest Task Success on both WorldGuide Bench and Video-CraftBench. Despite remaining execution and stopping errors, these results support a shift from open-loop future synthesis to closed-loop task execution. Our work takes a step towards how video world models can use their generated consequences to decide what to do next and when to stop, guiding step-by-step execution toward the final goal.

\bibliographystyle{plainnat}
\bibliography{main_ref}

\clearpage

\clearpage
\appendix

\newcommand{\bluelink}[2]{\hyperref[#1]{\textcolor{blue}{#2}}}
\definecolor{linkskyblue}{RGB}{30,144,255}
\newcommand{\hlurl}[1]{\href{#1}{\textcolor{linkskyblue}{\bfseries\urlstyle{same}\nolinkurl{#1}}}}

\begin{center}
    {\Large\bfseries Supplementary Contents}
\end{center}
\vspace{1em}
\hrule
\vspace{0.8em}
\noindent
\begin{minipage}{0.96\linewidth}
\small
\bluelink{app:design_rationale}{\textbf{\ref*{app:design_rationale}. Design Rationale}}\dotfill\pageref{app:design_rationale} \\[12pt]
\bluelink{app:dataset_details}{\textbf{\ref*{app:dataset_details}. WorldGuide Bench}}\dotfill\pageref{app:dataset_details} \\[6pt]
\hspace*{1.5em}\bluelink{app:dataset_collection}{\ref*{app:dataset_collection}. Collection and Filtering}\dotfill\pageref{app:dataset_collection} \\[5pt]
\hspace*{1.5em}\bluelink{app:dataset_gemini}{\ref*{app:dataset_gemini}. Temporal Action Annotation}\dotfill\pageref{app:dataset_gemini} \\[5pt]
\hspace*{1.5em}\bluelink{app:dataset_clips}{\ref*{app:dataset_clips}. Training Samples}\dotfill\pageref{app:dataset_clips} \\[5pt]
\hspace*{1.5em}\bluelink{app:dataset_stats}{\ref*{app:dataset_stats}. Statistics and Evaluation Split}\dotfill\pageref{app:dataset_stats} \\[5pt]
\hspace*{1.5em}\bluelink{app:dataset_example}{\ref*{app:dataset_example}. Annotation Example}\dotfill\pageref{app:dataset_example} \\[5pt]
\hspace*{1.5em}\bluelink{app:human_dataset_audit}{\ref*{app:human_dataset_audit}. Human Audit of Annotations}\dotfill\pageref{app:human_dataset_audit} \\[12pt]
\bluelink{app:memory_details}{\textbf{\ref*{app:memory_details}. Hierarchical Memory}}\dotfill\pageref{app:memory_details} \\[6pt]
\hspace*{1.5em}\bluelink{app:memory_bank}{\ref*{app:memory_bank}. History and Reference Conditioning}\dotfill\pageref{app:memory_bank} \\[5pt]
\hspace*{1.5em}\bluelink{app:memory_embedders}{\ref*{app:memory_embedders}. Multi-Scale Embedding and Injection}\dotfill\pageref{app:memory_embedders} \\[5pt]
\hspace*{1.5em}\bluelink{app:memory_budget}{\ref*{app:memory_budget}. Compression Schedule and Token Budget}\dotfill\pageref{app:memory_budget} \\[12pt]
\bluelink{app:training_details}{\textbf{\ref*{app:training_details}. Training and Inference}}\dotfill\pageref{app:training_details} \\[6pt]
\hspace*{1.5em}\bluelink{app:planner_training}{\ref*{app:planner_training}. ContextPlanner Training}\dotfill\pageref{app:planner_training} \\[5pt]
\hspace*{1.5em}\bluelink{app:executor_training}{\ref*{app:executor_training}. Executor Training}\dotfill\pageref{app:executor_training} \\[5pt]
\hspace*{1.5em}\bluelink{app:inference}{\ref*{app:inference}. Rollout and Termination}\dotfill\pageref{app:inference} \\[12pt]
\bluelink{app:eval_metrics}{\textbf{\ref*{app:eval_metrics}. Evaluation Protocol and Metrics}}\dotfill\pageref{app:eval_metrics} \\[6pt]
\hspace*{1.5em}\bluelink{app:eval_protocol}{\ref*{app:eval_protocol}. Benchmarks, Baselines, and Ablations}\dotfill\pageref{app:eval_protocol} \\[5pt]
\hspace*{1.5em}\bluelink{app:gemini_evaluation}{\ref*{app:gemini_evaluation}. Video-Judge Metrics}\dotfill\pageref{app:gemini_evaluation} \\[5pt]
\hspace*{1.5em}\bluelink{app:planner_evaluation}{\ref*{app:planner_evaluation}. Planner Evaluation}\dotfill\pageref{app:planner_evaluation} \\[5pt]
\hspace*{1.5em}\bluelink{app:alignment_metrics}{\ref*{app:alignment_metrics}. Paired Alignment}\dotfill\pageref{app:alignment_metrics} \\[5pt]
\hspace*{1.5em}\bluelink{app:vbench-evaluation}{\ref*{app:vbench-evaluation}. VBench Video-Quality Metrics}\dotfill\pageref{app:vbench-evaluation} \\[12pt]
\bluelink{app:human_model_study}{\textbf{\ref*{app:human_model_study}. Human Evaluation of Generated Videos}}\dotfill\pageref{app:human_model_study} \\[12pt]
\bluelink{app:additional_results}{\textbf{\ref*{app:additional_results}. Additional Results}}\dotfill\pageref{app:additional_results} \\[6pt]
\hspace*{1.5em}\bluelink{app:additional_qualitative}{\ref*{app:additional_qualitative}. WorldGuide Bench Qualitative Comparisons}\dotfill\pageref{app:additional_qualitative} \\[5pt]
\hspace*{1.5em}\bluelink{app:videocraft_results}{\ref*{app:videocraft_results}. Video-CraftBench Results}\dotfill\pageref{app:videocraft_results} \\[12pt]
\bluelink{app:failure_analysis}{\textbf{\ref*{app:failure_analysis}. Limitations and Failure Analysis}}\dotfill\pageref{app:failure_analysis}
\end{minipage}
\vspace{0.8em}
\hrule

\vspace{1.2em}
\paragraph{Interactive qualitative results:}
Our project page provides complete rollouts of WorldGuide and the baselines side by side, together with additional WorldGuide Bench videos and their step-level annotations: \hlurl{https://mbzuai-oryx.github.io/WorldGuide}. These pages are supplementary qualitative material; all quantitative results and conclusions are reported in the paper.

\clearpage
\section{Design Rationale}
\label{app:design_rationale}

This section explains three design choices of WorldGuide: the choice of backbones, the sequential training of the two modules, and the language-level interface between planning and execution.

\paragraph{One model for planning and action encoding.}
HunyuanVideo-1.5~\citep{wu2025hunyuanvideo} encodes its text prompts with Qwen2.5-VL-7B-Instruct~\citep{bai2025qwen25vltechnicalreport}. We therefore initialize the ContextPlanner from this same model, so that a single network both predicts the next action and encodes it for the Executor. This choice has three advantages. First, \emph{the Executor receives action embeddings from the model it was pretrained with, so fine-tuning adapts an existing interface rather than learning a new one}. Second, \emph{WorldGuide adds no separate planning network: the ContextPlanner takes the place of the Executor's text encoder, keeping the complete system at the size of the original HunyuanVideo-1.5 pipeline}. Third, \emph{the action predicted by the planner at inference reaches the Executor through exactly the embedding pathway used during training.}

\paragraph{Choice of Executor.}
HunyuanVideo-1.5 has 8.3B parameters, fewer than many of the evaluated generators, yet it is among the strongest at procedural execution. Given reference actions on WorldGuide Bench, it achieves the highest Plan Accuracy (43.53) and Task Success (12.00\%) of all baselines except the 33B MiniMax-H3, ahead of larger models such as Wan2.2-14B, LTX-2.5-22B, and FlashMotion-34B (Table~\ref{tab:main_results}). Building on it keeps WorldGuide compact while starting from a strong executor.

\paragraph{Sequential training.}
Both modules are trained on the same procedural demonstrations, but sequentially rather than end-to-end. The ContextPlanner is first fine-tuned for next-action and completion prediction; it is then frozen while the Executor is fine-tuned on action captions embedded by it (Appendix~\ref{app:training_details}). Freezing is deliberate. If the Executor's flow-matching loss also updated the ContextPlanner, its parameters would be optimized to produce embeddings that ease video generation, with no signal that preserves its language outputs. Its predicted actions could then drift away from faithful, readable instructions. Freezing the ContextPlanner keeps the interface between planning and execution in natural language.

\paragraph{Why a language-level interface matters.}
WorldGuide represents each predicted action as a natural-language instruction and uses it to generate the corresponding video clip. This language interface has three main benefits:
\begin{itemize}
    \item \textbf{Interpretable guidance.} Each generated clip is paired with a clear instruction describing the intended action. This makes the generated procedure easier to understand and follow.
    
    \item \textbf{Diagnosable failures.} Planning and execution errors can be analyzed separately. Since actions are represented as text, we can replace predicted actions with reference actions while keeping the Executor unchanged. We use this setting for the oracle ContextPlanner experiment in Table~\ref{tab:ablation_worldguide}.
    
    \item \textbf{Independent evaluation.} The predicted action sequence can be evaluated directly as text, independently of the quality of the generated videos (Table~\ref{tab:ablation_planner}).
\end{itemize}
In this work, we keep the language interface fixed during Executor training. Jointly optimizing the Planner and Executor while preserving clear language outputs is an interesting direction for future work (Sec.~\ref{sec:discussion}).

\vspace{2em}
\section{WorldGuide Bench}
\label{app:dataset_details}

WorldGuide Bench supplies the three signals on which the closed loop is trained: which atomic action comes next, how that action changes the visual state, and when the task is complete. This section describes how videos are collected (Appendix~\ref{app:dataset_collection}), annotated (Appendix~\ref{app:dataset_gemini}), converted into training samples (Appendix~\ref{app:dataset_clips}), and split (Appendix~\ref{app:dataset_stats}), followed by an annotation example (Appendix~\ref{app:dataset_example}) and a human audit (Appendix~\ref{app:human_dataset_audit}).

Table~\ref{tab:dataset_comparison} compares WorldGuide Bench with existing procedural video datasets. These target planning question answering (EgoPlan-IT~\citep{qiu2026egoplan}), step localization and grounding (Ego4D Goal-Step~\citep{song2023ego4d}, COIN~\citep{tang2019coin}, CrossTask~\citep{zhukov2019crosstask}, HT-Step~\citep{afouras2023ht}), segmentation, recognition, and error detection (YouCook2~\citep{zhou2018youcook2}, Assembly101~\citep{sener2022assembly101}, CaptainCook4D~\citep{peddi2024captaincook4d}, Ego-Exo4D Keystep~\citep{grauman2024ego}), or video generation and its evaluation (SemComp-Data~\citep{tu2026semcomp}, ActVideoGen~\citep{yang2026spiral}, EgoForge~\citep{shen2026egoforge}). None of them provides explicit task-completion supervision, which WorldGuide Bench adds to step-level atomic-action clips and task goals.

\subsection{Collection and Filtering}
\label{app:dataset_collection}

We collect public instructional videos of at most seven minutes from YouTube, covering 245 tasks in 27 procedural categories such as origami, cooking, construction, assembly and repair, sorting, and knotting. Each task is retrieved with task-specific queries expanded with short-form and multilingual variants in 15 languages, so every task has many demonstrations with different objects, scenes, and demonstrators. Files that fail integrity or frame-decoding checks are discarded.

Gemini~2.5~Flash~\citep{gemini25} rates each video with a Video Planning Score from 0 to 5 for topical relevance and demonstration completeness (Fig.~\ref{fig:gemini_caption_prompt}). We keep only videos scored 5, retaining 58,679 of 133,360 retrieved records (44.0\%; Table~\ref{tab:dataset_stats}). Annotation quality is verified by a human audit (Appendix~\ref{app:human_dataset_audit}).

All source videos are publicly available and used for non-commercial research. We will release source identifiers, timestamps, and step annotations rather than the videos themselves.

\subsection{Temporal Action Annotation}
\label{app:dataset_gemini}

Gemini converts each video into an ordered sequence of atomic physical actions. Each action is assigned a start and end timestamp, with a target duration of 1--5 seconds (Fig.~\ref{fig:gemini_caption_prompt}). Actions longer than 5 seconds are split into multiple consecutive steps. The final step is marked as task completion.
For videos longer than 5 minutes, we run annotation at $1.5\times$ playback speed for efficiency and rescale the predicted timestamps back to the original video timeline. As a result, annotated steps can span 1.5--7.5 seconds in the original video. All clips are extracted from the original-speed videos, and empty or invalid clips are removed.

\begin{figure*}[htbp]
\centering
\begin{tcolorbox}[
    width=0.96\textwidth,
    colback=white,
    colframe=black!55,
    colbacktitle=blue!7,
    coltitle=black,
    title=\textbf{Gemini Dataset-Captioning Prompt},
    fonttitle=\small\bfseries,
    boxrule=0.5pt,
    arc=2mm,
    left=7pt,
    right=7pt,
    top=5pt,
    bottom=5pt
]
\small
\ttfamily
You are an expert temporal video captioning model.\\
Task: Watch the video and convert it into a timeline of actionable steps. Temporal alignment between each step and its timestamp is very important.\\[2pt]
Main task/topic: \{query\}.\\
Description: \{description\}.\\[2pt]
Video Planning Score:\\
- 0: Video does not match the topic.\\
- 1-2: Poor quality, missing steps, or hard to follow.\\
- 3-4: Good quality, most steps visible.\\
- 5: Excellent, complete step-by-step tutorial.\\[2pt]
Timeline Rules:\\
- Output only the steps necessary to complete the task. Please align the tasks with the particular timeline.\\
- Each step must be between 1 and 5 seconds. (Adjusted for long-video stability).\\
- If an action takes longer than 5s, split it into consecutive steps.\\
- Use precise verbs: Fold, Aligns, Presses, Flips, Rotates.\\
- No descriptions of background, hands, or music.\\[2pt]
Important:\\
- If Score = 0, output "Video Planning Score: 0" and stop.\\
- If Score \textgreater{} 0, output the Score followed by the timeline.\\[2pt]
Strict output format:\\
Video Planning Score: \textless{}0-5\textgreater{}\\
Work subject: \textless{}one-line title describing the specific \{task\} of \{category\} shown in the video\textgreater{}\\
Time: \textless{}start\textgreater{}-\textless{}end\textgreater{} sec, Step \textless{}N\textgreater{}: \textless{}one clear physical action\textgreater{}
\end{tcolorbox}
\caption{Per-video prompt supplied to Gemini~2.5~Flash for dataset captioning. The video, together with the retrieval query, task, category, and description, is passed alongside this prompt, prefixed by a fixed system instruction: \emph{``You are a professional technical writer. Your task is to provide a non-repetitive, high-accuracy timeline of physical actions from video.''} Only videos assigned a Video Planning Score of 5 are retained.}
\label{fig:gemini_caption_prompt}
\end{figure*}

\subsection{Training Samples}
\label{app:dataset_clips}

Each annotated step becomes one training record (Table~\ref{tab:dataset_sample_format}): the task context, the action caption, the target clip cut at the step's timestamps, the step index and normalized progress, the preceding clips and captions, and a completion flag set on the final step. A one-second preview from the start of the source video provides the initial visual state. From these records, the ContextPlanner learns to predict the next caption from the goal and up to $K{=}3$ preceding clip--caption pairs, and the Executor learns to generate each clip from its caption and preceding visual context. To supervise stopping, preprocessing appends one additional record after the final step, whose target is \texttt{<|Task Completed|>} and whose history is the last $K{=}3$ clip--caption pairs, ending with the final step. This record has no target clip, so it trains only the ContextPlanner and matches inference, where the planner emits \texttt{<|Task Completed|>} after observing the last generated clip. Both modules are thus supervised by the same demonstrations, each with its own objective (Appendix~\ref{app:design_rationale}).

\begin{table}[htbp]
\centering
\small
\caption{\textbf{WorldGuide Bench sample format.} 
Each record contains an annotated action, its corresponding video clip, 
and the preceding procedural context. The completion flag identifies the final step.}
\label{tab:dataset_sample_format}

\renewcommand{\arraystretch}{1.25} 
\begin{tabular}{@{}p{0.22\linewidth}p{0.73\linewidth}@{}}
\toprule
\textbf{Information} & \textbf{Contents} \\
\midrule

\textbf{Task context} 
& Category, task label, and task goal. \\

\textbf{Action annotation} 
& Step index, action caption, and start/end timestamps in the original video. \\

\textbf{Target clip} 
& Video segment corresponding to the annotated action. \\

\textbf{Preceding history} 
& Previous clips and their action captions. The ContextPlanner uses up to $K=3$ previous steps, while the Executor uses visual memory. \\

\textbf{Progress} 
& Step index normalized by the total number of annotated steps. \\

\textbf{Completion} 
& Flag marking the final step. After it, one ContextPlanner-only record with target \texttt{<|Task Completed|>} and no target clip is appended. \\

\bottomrule
\end{tabular}
\end{table}

\subsection{Statistics and Evaluation Split}
\label{app:dataset_stats}

Table~\ref{tab:dataset_stats} reports the number of videos after each filtering stage. We hold out four videos per task, giving 980 test videos and 57,699 training videos; all clips of a held-out video are excluded from the training of both modules. The split therefore evaluates unseen demonstrations of known tasks.



\subsection{Annotation Example}
\label{app:dataset_example}

Table~\ref{tab:dataset_annotation_example} shows a complete annotation example from an origami video. Each video in WorldGuide Bench follows the same format: a task description, a short description of the work subject, and an ordered sequence of atomic actions with their start and end times. During preprocessing, we split the video into one clip for each annotated step. We then organize the clips, action captions, step order, and completion flag into the step-level records described in Appendix~\ref{app:dataset_clips}. The example in Table~\ref{tab:dataset_annotation_example} contains 21 aligned clip--caption pairs. Step~21 carries the completion flag, so the ContextPlanner's target after it is \texttt{<|Task Completed|>}. More annotation examples are available on the project page.

\begin{wraptable}{r}{0.48\textwidth}
\centering
\small
\caption{\textbf{WorldGuide Bench construction statistics.} Counts are reported after each filtering stage. Only videos receiving a Gemini Video Planning Score of 5/5 are retained. The video-disjoint test split holds out four videos per task.}
\label{tab:dataset_stats}
\resizebox{\linewidth}{!}{%
\begin{tabular}{lr}
\toprule
\textbf{Quantity} & \textbf{Count} \\
\midrule
Top-level categories & 27 \\
Task topics & 245 \\
Merged raw video metadata records & 133,360 \\
Validated (decodable) video records & 132,911 \\
Videos with a non-null Gemini annotation & 119,798 \\
Videos with planning score $=$ 5 (retained) & 58,679 \\
\midrule
Training split videos & 57,699 \\
Test split videos (held out) & 980 \\
\quad videos per task & 4 \\
\quad tasks / categories & 245 / 27 \\
\bottomrule
\end{tabular}%
}
\end{wraptable}

This example shows three important properties of our annotations. \textit{First}, each caption describes a single physical action. This includes fine-grained intermediate actions, such as folding and then unfolding the paper in Steps 1--4. \textit{Second}, the same action can appear at different stages of a task. For example, \textit{``Unfolds the paper''} appears at Steps 2, 4, and 7, while \textit{``Flips the paper over''} appears at Steps 5, 8, 16, and 19. Similarly, \textit{``Presses the wing fold''} appears at Steps 18 and 21. Therefore, the action text alone is not enough to determine the current task progress. The ContextPlanner also observes the generated clips to understand the current visual state (Sec.~\ref{sec:visual_feedback}). \textit{Third}, the \texttt{<|Task Completed|>} target after the final step provides supervision for learning when to stop.

For each step, the Executor learns to generate the corresponding video clip from the action caption and previous visual context. The ContextPlanner learns to predict the next action from up to $K{=}3$ previous clip-action pairs. For example, Steps 15--17 are used as context when predicting Step 18, and Steps 19--21 are used as context when predicting \texttt{<|Task Completed|>}.

\begin{table}[t]
\centering
\small
\caption{\textbf{Complete annotation of one WorldGuide Bench video.}
Task: origami, \textit{airplane\_dart}. Work subject: folding a paper airplane (dart style). Each row is one clip; Time gives its start and end in the source video (mm:ss).}
\label{tab:dataset_annotation_example}
\begin{tabular}{@{}ccp{0.70\linewidth}@{}}
\toprule
\textbf{Step} & \textbf{Time} & \textbf{Action caption} \\
\midrule
1 & 00:04--00:05 & Folds the bottom right corner of the paper to the top left. \\
2 & 00:05--00:06 & Unfolds the paper. \\
3 & 00:06--00:07 & Folds the bottom left corner of the paper to the top right. \\
4 & 00:07--00:08 & Unfolds the paper. \\
5 & 00:08--00:09 & Flips the paper over. \\
6 & 00:09--00:10 & Folds the bottom edge of the paper upwards. \\
7 & 00:10--00:11 & Unfolds the paper. \\
8 & 00:11--00:12 & Flips the paper over. \\
9 & 00:12--00:13 & Collapses the side edges inwards to form a triangle at the bottom. \\
10 & 00:13--00:14 & Presses the folded triangle flat. \\
11 & 00:14--00:15 & Folds the left corner of the triangle towards the center. \\
12 & 00:15--00:16 & Folds the right corner of the triangle towards the center. \\
13 & 00:16--00:17 & Folds the bottom tip of the triangle upwards. \\
14 & 00:17--00:18 & Presses the fold firmly. \\
15 & 00:18--00:19 & Folds the entire paper in half lengthwise. \\
16 & 00:19--00:20 & Flips the paper over. \\
17 & 00:20--00:21 & Folds down one wing, aligning it with the bottom edge. \\
18 & 00:21--00:22 & Presses the wing fold. \\
19 & 00:22--00:23 & Flips the paper over. \\
20 & 00:23--00:24 & Folds down the second wing, aligning it with the bottom edge. \\
21 & 00:24--00:25 & Presses the wing fold. \\
\midrule
-- & -- & \texttt{<|Task Completed|>} (ContextPlanner target after Step 21; no clip) \\
\bottomrule
\end{tabular}
\end{table}

\subsection{Human Audit of Annotations}
\label{app:human_dataset_audit}

Since WorldGuide Bench is automatically annotated, we conduct a human audit to evaluate the annotation quality. Three independent annotators review 245 videos, with one video sampled from each task. These videos contain a total of 3,168 annotated action steps. For each video, the annotators watch the original video together with its task label and ordered action captions. Each annotator evaluates the annotations independently.

\paragraph{Evaluation criteria.}
We evaluate the annotations using six criteria. Each criterion is answered based only on the visual evidence in the video.
\begin{itemize}
    \item \textbf{Caption correctness} (per step): Does the caption correctly describe the visible action, the involved objects, and the result of the action?
    \item \textbf{Action atomicity} (per step): Does the caption describe one clear action rather than multiple actions?
    \item \textbf{Task-label correctness} (per video): Does the task label correctly describe the task performed in the video?
    \item \textbf{Essential-action coverage} (per video): Do the captions cover all actions needed to complete the task?
    \item \textbf{Temporal order} (per video): Are the captions ordered according to when the actions occur in the video?
    \item \textbf{Demonstration completeness} (per video): Does the final visible state show that the task has been completed?
\end{itemize}

\paragraph{Scoring.}
For each item, annotators choose \emph{Correct} ($1$), \emph{Incorrect} ($0$), or \emph{Cannot determine} ($U$). The $U$ option is used when the video does not provide enough visual evidence to make a decision, and these ratings are excluded from the score. The score is the percentage of rated items marked as Correct. For step-level criteria, scores are first averaged within each video and then across videos, so that every task contributes equally. We also report a majority-vote score. An item receives a final Correct or Incorrect label when at least two of the three annotators agree.

\begin{wraptable}{r}{0.57\textwidth}
    \vspace{-1em}
    \centering
    \small
    \setlength{\tabcolsep}{4pt}
    \caption{\textbf{Human audit of WorldGuide Bench annotations} (\% judged correct) on 245 task-balanced videos with 3,168 action steps. \textit{Majority} aggregates the three annotators by item-level vote.}
    \label{tab:audit_majority}
    \resizebox{\linewidth}{!}{%
    \begin{tabular}{@{}llcccc@{}}
        \toprule
        Criterion & Level & Annot.~1 & Annot.~2 & Annot.~3 & Majority \\
        \midrule
        Caption correctness        & Step  & 82.4 & 82.6 & 81.9 & \textbf{83.0} \\
        Action atomicity           & Step  & 82.2 & 81.5 & 82.3 & \textbf{82.9} \\
        Task-label correctness     & Video & 82.9 & 82.4 & 83.2 & \textbf{83.7} \\
        Essential-action coverage  & Video & 83.2 & 84.4 & 83.9 & \textbf{84.5} \\
        Temporal order             & Video & 82.1 & 81.0 & 81.6 & \textbf{81.6} \\
        Demonstration completeness & Video & 81.4 & 81.7 & 82.6 & \textbf{82.9} \\
        \bottomrule
    \end{tabular}%
    }
    \vspace{-2em}
\end{wraptable}

\paragraph{Results.}
Table~\ref{tab:audit_majority} summarizes the human audit results. The majority-vote scores range from 81.6\% to 84.5\% across all six criteria. Caption correctness reaches 83.0\%, task-label correctness reaches 83.7\%, and essential-action coverage reaches 84.5\%. The scores from the three annotators differ by at most 1.3 percentage points for each criterion, showing similar judgments across annotators. Overall, the audit indicates that the automatic annotation pipeline provides reliable supervision for action captions, temporal order, and task completion.


\vspace{2em}
\section{Hierarchical Memory}
\label{app:memory_details}

The Executor uses the spatial history compression strategy from YUME~\citep{mao2025yume,zhang2026frame}. Recent latent frames are kept at higher spatial resolution, while older frames are progressively compressed to lower resolutions. This keeps the number of memory tokens bounded as the generated video becomes longer.

\subsection{History and Reference Conditioning}
\label{app:memory_bank}

\paragraph{Terminology.}
The history contains $F\leq T+1\leq1366$ latent frames. It includes up to $T\leq1365$ recent frames and the oldest frame, which we call the \emph{anchor}. We also use a separate \emph{reference} slot containing the latest generated frame, i.e., the last frame of the current visual state $x_t$ in Eq.~\ref{eq:rollout} (the initial image $x_0$ at the first step). The reference may contain the same frame as the newest history entry, but it uses separate token positions.

We use $L$ to denote the latent length of the clip being generated. All frame counts in this section refer to latent frames after VAE encoding. For $N$ continuously encoded video frames, the VAE produces
$\lfloor(N-1)/4\rfloor+1$ latent frames.

\paragraph{Supplied history.}
To construct the history, we concatenate the latent frames from previous clips and keep the most recent frames that fit within the available history length. If the history is shorter than the required length, we left-pad it; these padded positions add tokens but contain no visual information. Since one conditioning slot is reserved for the reference frame, the history length satisfies
\[
F\leq\min(L-1,T+1).
\]
The history is updated after each generated clip. If there is no previous history, the memory branch is skipped.

\paragraph{Conditioning input.}
The input to the DiT is
\begin{equation}
x_{\mathrm{DiT}}
=
[\,\tilde z\,;\,z_{\mathrm{cond}}\,;\,m_{\mathrm{cond}}\,]
\in
\mathbb{R}^{(2C+1)\times L\times H\times W}.
\end{equation}
Here, $\tilde z$ is the noisy video latent, $z_{\mathrm{cond}}$ contains the visual conditioning, and $m_{\mathrm{cond}}$ indicates which conditioning positions are valid. Without memory, only the reference slot in $z_{\mathrm{cond}}$ is filled. With memory, $z_{\mathrm{cond}}$ contains the history followed by the reference frame. The same history latents are also spatially compressed into memory tokens, as described in Appendix~\ref{app:memory_embedders}.

\subsection{Multi-Scale Embedding and Injection}
\label{app:memory_embedders}

\paragraph{Multi-scale embedders.}
We use different spatial resolutions to represent the visual history. With patch size $p=2$, we define embedders $E_r$ for compression rates $r\in\{1,2,4,8,16\}$. Each $E_r$ is a 3-D convolution with kernel size and stride $(1,pr,pr)$, which maps $C$ input channels to $d$-dimensional features. $E_1$ uses the pretrained patch-embedding weights. For larger compression rates, we initialize the convolution kernels by trilinearly interpolating the pretrained weights.

For the coarsest rate, $r=64$, we first apply a channel-preserving 3-D convolution with kernel size and stride $(1,4,4)$, followed by $E_{16}$. This gives an effective spatial stride of $128$. The temporal kernel size and stride are always one. Therefore, the compression only reduces the number of spatial tokens while preserving all temporal latent frames.

\paragraph{Positional encoding and memory injection.}
Each compressed history segment receives 3-D rotary positional embeddings based on its spatial grid. Temporal positions continue across consecutive history segments, preserving their temporal order. These positions correspond to latent-frame positions rather than timestamps in the original video.

The resulting memory tokens $\mathcal{M}$ are added to the Executor's context sequence together with the visual and text conditioning:
\begin{equation}
\tau =
[\,\mathcal{M}\,;\,c_{\text{vision}}\,;\,c_{\text{text}}\,].
\end{equation}
This adds visual-history information as conditioning tokens without increasing the number of video tokens being denoised or changing the training objective in Eq.~\ref{eq:exec}. When no history is available, the memory branch is skipped and the Executor uses only the reference-image conditioning.

\subsection{Compression Schedule and Token Budget}
\label{app:memory_budget}

The schedule is selected by the history length $F$. Table~\ref{tab:memory_budget} gives the longest schedule ($343\leq F\leq1366$), with ages running from the newest frame ($1$) to the anchor ($F$).


\paragraph{Shorter histories.}
For $7\leq F\leq342$, the newest three frames use rate one, the next two use rate two, and successively older tiers use rates four, eight, and sixteen as needed. The oldest anchor uses rate one for $F\leq86$ and rate two thereafter. For $F\leq6$, both endpoints use rate one and the intervening frames use rate two. For $F\leq2$, the empty intermediate slice reuses the newest latent, as in YUME. Using nominal area ratios before spatial padding, the bounds including anchor and reference are
\begin{equation}
\frac{p^2}{HW}|\mathcal{M}|_{\mathrm{nominal}}\leq
\begin{cases}
4, & 1\leq F\leq6,\\
6.5, & 7\leq F\leq22,\\
7.5, & 23\leq F\leq86,\\
7.75, & 87\leq F\leq342,\\
8, & 343\leq F\leq1366.
\end{cases}
\end{equation}
For every history length, the history, anchor, and reference therefore cost at most eight full-resolution latent-frame equivalents before spatial padding. This bounds the number of context tokens, not the number of retained temporal frames.

\begin{wraptable}{r}{0.5\textwidth}
\vspace{-1em}
\centering
\small
\setlength{\tabcolsep}{4pt}
\caption{Nominal token cost for $343\leq F\leq1366$ history latent frames. One full-resolution latent-frame equivalent is $HW/p^2$ tokens; the reference is counted separately.}
\label{tab:memory_budget}
\resizebox{\linewidth}{!}{%
\begin{tabular}{lcrr}
\toprule
Segment & Latent frames & Rate & Cost upper bound \\
\midrule
Present & $3$ & 1 & 3.00 \\
Near present & $2$ & 2 & 0.50 \\
Recent past & $16$ & 4 & 1.00 \\
Mid past & $64$ & 8 & 1.00 \\
Far past & $256$ & 16 & 1.00 \\
Distant past & $F-342\leq1024$ & 64 & 0.25 \\
Oldest anchor & $1$ & 2 & 0.25 \\
\midrule
History subtotal & $F$ & -- & 7.00 \\
Reference & $1$ & 1 & 1.00 \\
\midrule
\textbf{Total} & $F+1$ & -- & \textbf{8.00} \\
\bottomrule
\end{tabular}%
}
\vspace{-2em}
\end{wraptable}

\paragraph{Spatial padding.}
With padding, a rate-$r$ embedder produces $\lceil H/(pr)\rceil\lceil W/(pr)\rceil$ tokens per latent frame, the coarsest embedder $\lceil H/128\rceil\lceil W/128\rceil$, and the reference $\lceil H/p\rceil\lceil W/p\rceil$. These integer counts can exceed the nominal bound when $H$ and $W$ are not divisible by the strides, but for a fixed history capacity and resolution the token count remains bounded independently of rollout length.

\vspace{2em}
\section{Training and Inference}
\label{app:training_details}

The two modules are trained sequentially on the WorldGuide Bench training split (Appendix~\ref{app:dataset_stats}): first the ContextPlanner, then the Executor with the ContextPlanner frozen (Appendix~\ref{app:design_rationale}).

\subsection{ContextPlanner Training}
\label{app:planner_training}

We fine-tune all parameters of Qwen2.5-VL-7B-Instruct~\citep{bai2025qwen25vltechnicalreport} to predict the next action or the completion token from the task goal, the current visual state, and up to $K{=}3$ preceding clip--action pairs. The cross-entropy loss is applied only to response tokens (Table~\ref{tab:planner_training_details}).

\begin{table}[t]
\centering
\small
\caption{ContextPlanner training configuration. Full-parameter fine-tuning of Qwen2.5-VL-7B-Instruct for next-step action prediction from goal, visual state, and a history of $K{=}3$ clip--action pairs.}
\label{tab:planner_training_details}
\setlength{\tabcolsep}{4pt}
\begin{tabular}{@{}p{0.35\linewidth}p{0.59\linewidth}@{}}
\toprule
\textbf{Component} & \textbf{Configuration} \\
\midrule
Base model & Qwen2.5-VL-7B-Instruct \\
Training objective & Autoregressive next-step prediction \\
Trainable parameters & Full-parameter fine-tuning \\
Precision & BF16 \\
Distributed training & DeepSpeed ZeRO-3 \\
Hardware & 8 nodes $\times$ 8 AMD Instinct MI210 (64\,GB) \\
Training time & 7 days \\
Per-GPU batch size & 2 \\
Gradient accumulation & 8 \\
Effective global batch size & 1024 \\
Optimizer & AdamW \\
Learning rate & $2 \times 10^{-5}$ \\
Weight decay & 0.01 \\
LR schedule & Cosine \\
Warmup steps & 100 \\
Epochs & 2 \\
Maximum sequence length & 4096 \\
History length $K$ & 3 clip--action pairs \\
\bottomrule
\end{tabular}
\end{table}

\subsection{Executor Training}
\label{app:executor_training}

We fine-tune HunyuanVideo-1.5~\citep{wu2025hunyuanvideo} for action-conditioned image-to-video generation with the masked flow-matching objective of Eq.~\ref{eq:exec}, using precomputed VAE latents. Each action caption is embedded by the frozen ContextPlanner, and the multi-scale memory embedders are trained together with the Executor (Table~\ref{tab:executor_training_details}). The memory ablation evaluates this final checkpoint with and without memory (Appendix~\ref{app:eval_protocol}).

\begin{table}[t]
\centering
\small
\caption{Executor training configuration. HunyuanVideo-1.5 DiT fine-tuned for language-conditioned image-to-video generation on step-level clips under the masked flow-matching objective (Eq.~\ref{eq:exec}).}
\label{tab:executor_training_details}
\setlength{\tabcolsep}{4pt}
\begin{tabular}{@{}p{0.35\linewidth}p{0.59\linewidth}@{}}
\toprule
\textbf{Component} & \textbf{Configuration} \\
\midrule
Base model & HunyuanVideo-1.5 DiT \\
Training objective & Masked MSE flow matching \\
Input mode & Image-to-video \\
Target resolution & $480 \times 832$ \\
Target FPS & 24 \\
Text conditioning & Action-caption embeddings from the frozen ContextPlanner, and ByT5~\citep{xue2022byt5tokenfreefuturepretrained} features \\
Visual conditioning & Image-conditioned latent branch and vision states \\
Hardware & 8 nodes $\times$ 8 AMD Instinct MI210 (64\,GB) \\
Training time & 42 days \\
Learning rate & $1 \times 10^{-5}$ \\
Weight decay & $1 \times 10^{-4}$ \\
Gradient clipping & 1.0 \\
Precision & Mixed BF16; DiT backbone in FP32 \\
Diffusion scheduler & FlowMatchEulerDiscreteScheduler \\
Timestep sampling & Logit-normal \\
Logit-normal parameters & mean $0.0$, std.\ $1.0$ \\
CFG dropout rate & 0.1 \\
Distributed training & FSDP \\
Sequence parallelism & Disabled (size 1) \\
Per-GPU batch size & 1 \\
Gradient accumulation & 2 \\
Effective global batch size & 128 video segments \\
\bottomrule
\end{tabular}
\end{table}

\subsection{Rollout and Termination}
\label{app:inference}

At the first step, the ContextPlanner receives the task goal $g$ and the initial image $x_0$; afterwards it receives the latest generated clip $x_t$ and the history $h_t$ of the latest $K{=}3$ generated clip--action pairs (Eq.~\ref{eq:rollout}). When it emits \texttt{<|Task Completed|>}, the rollout stops immediately and no further clip is generated. Otherwise, the Executor renders the predicted action $a_t$ as the next clip $x_{t+1}$, and its memory is refreshed with this clip (Appendix~\ref{app:memory_bank}). Rollouts are capped at 80 steps, above the longest reference procedure in WorldGuide Bench (62 steps). The planner is never given the reference step count, and a rollout that reaches the cap is scored as generated rather than marked successful. The generated clips are concatenated into the final video. Because each clip is generated separately and the memory token count is bounded (Appendix~\ref{app:memory_budget}), the per-step cost does not grow with rollout length.

Sampling uses 10 denoising steps, guidance scale 7.5, and flow shift 5.0. Training clips are sampled at 24\,FPS, and generated videos are played back at 16\,FPS.

\vspace{2em}
\section{Evaluation Protocol and Metrics}
\label{app:eval_metrics}

\subsection{Benchmarks, Baselines, and Ablations}
\label{app:eval_protocol}

\paragraph{Baselines.}
We compare WorldGuide with recent video generation models~\citep{wu2025hunyuanvideo,wan2025wan,yuan2026helios,wei2026univideo,li2026flashmotion,zheng2025open,yang2025cogvideox,li2026skyreels}, video world models~\citep{ali2025worldcosmos25,sun2025worldplay,mao2025yume,fu2026learning_robomaster,wu2026video_spmem,tian2026astra,teng2025magi,hacohen2026ltx}, and three planner--executor methods: Bernini~\citep{team2026bernini}, PhysAgent~\citep{li2026physagent}, and TempAct~\citep{wang2026tempact}. These planner--executor methods use their own planning--execution loops and are marked with $*$ in the result tables.

\paragraph{WorldGuide Bench.}
We evaluate one rollout for each of the 980 held-out videos (Table~\ref{tab:main_results}). Unstarred baselines are given the complete ground-truth action plan. They generate the procedure step by step, following the reference actions in order. For each step, the model receives one action caption and continues generation from the last generated frame. Each step is generated for approximately 2 seconds (48 frames at 24\,FPS or 32 frames at 16\,FPS, depending on the lengths supported by each model). Therefore, these baselines know the full action sequence before generation begins.

In contrast, the starred planner--executor baselines and WorldGuide receive only the initial image and task goal. They must determine how to complete the task using their own planning process. No model receives future ground-truth video frames.

\paragraph{Video-CraftBench.}
We evaluate all 294 samples from Video-CraftBench\_v0.1~\citep{ren2026videoworld}, which contains block assembly and paper folding tasks.\footnote{\url{https://huggingface.co/maverickrzw/Video-CraftBench_v0.1}} In this benchmark, all models receive only the initial image and task goal, without a reference action plan (Table~\ref{tab:main_results2}). Since no reference plan is provided, the judge evaluates whether the generated video completes the sub-goals needed to achieve the given task. Different valid procedures and step granularities are accepted.

\paragraph{Ablation variants.}
Table~\ref{tab:ablation_worldguide} evaluates four variants using the same evaluation protocol.
\begin{itemize}
    \item \textbf{Open-loop:} The model predicts the full action sequence once from the initial image and task goal, then executes all actions without replanning. The action sequence and video length are therefore fixed before generation starts.

    \item \textbf{W/o visual:} The model still predicts actions step by step, but generated video clips are removed from the ContextPlanner input. The Executor remains unchanged.

    \item \textbf{W/o Mem:} We use the same final Executor checkpoint but disable its memory branch. Each new clip is conditioned only on the reference image.

    \item \textbf{Oracle ContextPlanner:} We replace the predicted actions with the ground-truth actions and provide them to the same Executor and memory in their reference order. Generation stops after the final reference action. The model does not receive any future ground-truth video frames.
\end{itemize}

\paragraph{Judges.}
We use Gemini~3.5~Flash~\citep{google2026gemini35flash} to evaluate the generated videos in Tables~\ref{tab:main_results}, \ref{tab:main_results2}, and~\ref{tab:ablation_worldguide}. We use GPT-5.2~\citep{openai2025gpt52} to evaluate the predicted text plans in Table~\ref{tab:ablation_planner}. Although the two judges share some metric names, they evaluate different aspects of the output, as summarized in Table~\ref{tab:metric_map}. Their scores should therefore not be compared directly. Judge responses that cannot be parsed are excluded, and each score is averaged over the remaining valid judgments. We further validate the video evaluation with the independent human study in Appendix~\ref{app:human_model_study}, where WorldGuide is ranked first, followed by MiniMax-H3.

\begin{table}[t]
\centering
\small
\caption{Metric meanings on WorldGuide Bench. Video scores assess rendered execution; text scores assess the predicted procedure. Video-CraftBench uses goal-based video judging without a reference plan.}
\label{tab:metric_map}
\setlength{\tabcolsep}{4pt}
\begin{tabular}{@{}p{0.18\linewidth}p{0.38\linewidth}p{0.38\linewidth}@{}}
\toprule
& \textbf{Video judge} & \textbf{Text judge} \\
\midrule
Input & Rollout, goal, reference plan & Predicted plan, goal, reference plan \\
Plan Accuracy & Visibly achieved sub-goals & Semantic correctness and coverage \\
Order Score & Executed prerequisite order & Planned prerequisite order \\
Repeat/Skip & Unnecessary executions; omissions reduce Plan Accuracy & Redundant actions and missing sub-goals \\
Success & Final visible task completion & Plan would succeed if executed correctly \\
\bottomrule
\end{tabular}
\end{table}

\subsection{Video-Judge Metrics}
\label{app:gemini_evaluation}

The video judge (\texttt{gemini-3.5-flash}, temperature $0$) receives the complete generated video, the task description, and, on WorldGuide Bench, the ordered reference plan with the completion token removed; it never sees the ground-truth video (Fig.~\ref{fig:gemini_task_prompt}). From its structured JSON output, we compute four metrics for video $i$ with $N_i$ reference steps. Equivalent actions, different step granularities, and valid reorderings of independent steps receive credit.

\begin{figure*}[t]
\centering
\begin{tcolorbox}[
    width=0.96\textwidth,
    colback=white,
    colframe=black!55,
    colbacktitle=blue!7,
    coltitle=black,
    title=\textbf{Gemini Task-Evaluation Prompt Structure},
    fonttitle=\small\bfseries,
    boxrule=0.5pt,
    arc=2mm,
    left=7pt,
    right=7pt,
    top=5pt,
    bottom=5pt
]
\small
\textbf{Role and objective:}\quad
Evaluate visible procedural execution against the task goal, using the
ground-truth plan as a reference for required sub-goals and dependencies.

\vspace{4pt}
\textbf{Inputs:}\quad complete generated video; task description; ordered
ground-truth steps $1,\ldots,N$.

\vspace{4pt}
\textbf{Rules:}\quad watch the complete video; accept semantically equivalent
actions, different step granularities, and valid reorderings of independent
steps; credit a sub-goal only when its required effect is visibly achieved;
complete every required field and return only valid JSON.

\vspace{4pt}
\textbf{Metric instructions:}
\begin{itemize}
    \item \textbf{Plan Accuracy:} determine whether each reference step's
    required sub-goal is visibly achieved, including by equivalent actions;
    report $100\times(\text{achieved sub-goals}/N)$.
    \item \textbf{Order Score:} among executed steps, report
    $100\times(\text{steps respecting prerequisites}/\text{executed steps})$; use zero
    when no step is executed.
    \item \textbf{Unnecessary Repeat:} count repeated executions that add no
    task progress; report $100\times(\text{unnecessary repetitions}/N)$.
    Omissions reduce Plan Accuracy; valid changes in granularity are not repetitions.
    \item \textbf{Task Success:} return $1$ only if the final visible state
    shows that the complete task is achieved; otherwise return $0$.
\end{itemize}

\textbf{Required JSON structure:}\quad per-step execution/correctness and
reasons; observed step sequence and ordering violations; repeated steps and
extra counts; binary task success, final-state description, supporting and
opposing evidence; and a scalar summary of all four metrics.
\end{tcolorbox}
\caption{Prompt structure for procedural video evaluation with the Gemini~3.5~Flash judge (temperature $0.0$), used to produce the video-judge columns of Table~\ref{tab:main_results}. The judge receives the generated rollout, task description, and ground-truth plan with the \texttt{<|Task Completed|>} token stripped, and scores Plan Accuracy, Order Score, Repeat/Skip, and Task Success from visible execution alone; the ground-truth video is never shown.}
\label{fig:gemini_task_prompt}
\end{figure*}

\paragraph{Plan Accuracy ($\uparrow$).}
With $c_{ij}=1$ if the sub-goal of step $j$ is visibly achieved and $0$ otherwise,
\begin{equation}
    \mathrm{PlanAcc}_i = \frac{100}{N_i}\sum_{j=1}^{N_i}c_{ij}.
\end{equation}
Missing, partial, or ambiguous effects receive no credit.

\paragraph{Order Score ($\uparrow$).}
Of the $N_i^{\mathrm{executed}}$ observed steps, $N_i^{\mathrm{ordered}}$ respect their prerequisites:
\begin{equation}
    \mathrm{Order}_i = 100\,\frac{N_i^{\mathrm{ordered}}}{N_i^{\mathrm{executed}}},
\end{equation}
and $\mathrm{Order}_i=0$ if no step is executed. Because a short but correctly ordered execution can score highly, Order is read together with Plan Accuracy.

\paragraph{Repeat/Skip ($\downarrow$).}
With $R_i$ repeated executions that add no task progress,
\begin{equation}
    \mathrm{Repeat}_i = 100\,\frac{R_i}{N_i}.
\end{equation}
For the video judge, omitted steps are penalized by Plan Accuracy rather than by this term.

\paragraph{Task Success Rate ($\uparrow$).}
$\mathrm{Success}_i=1$ only if the final visible state achieves the complete task. Over $N_{\mathrm{eval}}$ valid judgments,
\begin{equation}
    \mathrm{TSR} = \frac{100}{N_{\mathrm{eval}}}\sum_{i=1}^{N_{\mathrm{eval}}}\mathrm{Success}_i.
\end{equation}
Plan Accuracy, Order, and Repeat/Skip are averaged over videos in the same way.

\paragraph{Skipped judge responses.}
For a few videos, the Gemini judge returns a response that cannot be parsed into the required JSON format. We skip these videos, so Task Success is computed over the remaining $N_{\mathrm{eval}}$ videos rather than all $980$. For example, on WorldGuide Bench: WorldGuide $326/978=33.33\%$ (2 skipped due to format issue in Gemini judge limitation), MiniMax-H3 $293/980=29.90\%$ (0 skipped), HunyuanVideo-1.5 $117/975=12.00\%$ (5 skipped), Cosmos-Predict2.5 $117/980=11.94\%$ (0 skipped), Bernini $19/971=1.96\%$ (9 skipped), and HY-WorldPlay $12/956=1.26\%$ (24 skipped, the maximum across models).

\subsection{Planner Evaluation}
\label{app:planner_evaluation}

Table~\ref{tab:ablation_planner} evaluates the action sequences predicted by the planner as text. The upper block evaluates Qwen2.5-VL and the ContextPlanner with different history lengths ($K\in\{1,2,3\}$). At each step, the planner predicts the next action from the task goal and the ground-truth clip--action history. The lower block evaluates the full WorldGuide loop, where the model runs autonomously using its own generated clips and predicted actions. The \emph{w/o visual} variant follows the same loop but removes generated clips from the planner input.

We use GPT-5.2~\citep{openai2025gpt52} to evaluate the predicted action sequences (Fig.~\ref{fig:gpt_planner_eval}). The judge receives the task description, predicted action sequence, and reference plan, but does not see the video. The reference plan defines the required sub-goals and their dependencies, but it is not treated as the only correct solution. The judge accepts equivalent actions, different step granularities, and valid changes in action order.
We report four metrics. \textbf{Plan Accuracy} measures whether the predicted actions are semantically correct and cover the required sub-goals. \textbf{Order Score} measures whether prerequisite actions appear in the correct order. \textbf{Repeat/Skip} measures redundant actions and missing sub-goals. \textbf{Plan Success} measures whether the full predicted plan would achieve the task goal if all actions were executed correctly.
Plan Success evaluates the action sequence itself, not whether the generated video stops at the correct visual state. Visual stopping is evaluated separately by the video judge. Stopping before the task is complete leads to failure in Task Success, while continuing after completion typically introduces unnecessary repeated actions.

\begin{figure*}[t]
\centering
\begin{tcolorbox}[
    width=0.96\textwidth,
    colback=white,
    colframe=black!55,
    colbacktitle=blue!7,
    coltitle=black,
    title=\textbf{GPT-5.2 Planner-Evaluation Prompt Structure},
    fonttitle=\small\bfseries,
    boxrule=0.5pt,
    arc=2mm,
    left=7pt,
    right=7pt,
    top=5pt,
    bottom=5pt
]
\small

\textbf{Role and objective:}
You are an expert evaluator of procedural task plans.
Determine whether the predicted action sequence provides a correct and
complete procedure for accomplishing the given task.

\vspace{4pt}
\textbf{Inputs:}
task description; predicted plan; ordered ground-truth reference plan.

\vspace{4pt}
\textbf{Reference-plan interpretation:}
The ground-truth plan describes one valid procedure and is provided to clarify
the task, required sub-goals, and action dependencies. Do not require exact
step-by-step matching. Accept semantically equivalent actions, different step
granularities, and alternative valid procedures that achieve the same task.

\vspace{4pt}
\textbf{Evaluation rules:}
\begin{itemize}
    \item Evaluate the complete predicted plan before assigning scores.
    \item Judge semantic and procedural correctness, not lexical similarity.
    \item Do not penalize valid reordering of independent actions.
    \item A reference action is not considered skipped if its required effect
          is achieved through another valid action.
    \item Penalize missing prerequisites, invalid ordering, redundant actions,
          and actions that do not contribute to task completion.
\end{itemize}

\vspace{2pt}
\textbf{Metrics:}
\begin{itemize}
    \item \textbf{Plan Accuracy:} correctness and completeness of the predicted
          procedure with respect to the task objective.
    \item \textbf{Order Score:} correctness of procedural dependencies and
          prerequisite ordering.
    \item \textbf{Repeat/Skip:} unnecessary repeated actions and uncovered
          task-critical sub-goals.
    \item \textbf{Plan Success:} whether executing the complete predicted plan
          correctly would accomplish the task.
\end{itemize}

\end{tcolorbox}
\caption{Prompt structure for text-only planner evaluation with a GPT-5.2 judge, used to produce Table~\ref{tab:ablation_planner}. The judge receives the task description, predicted plan, and ground-truth plan, with no video, and scores the procedural validity of the predicted text. The ground truth serves as a reference rather than a unique target, so semantically equivalent actions, orderings, and granularities are accepted.}
\label{fig:gpt_planner_eval}
\end{figure*}

\subsection{Paired Alignment}
\label{app:alignment_metrics}

Alignment compares each generated video with its paired ground-truth video using deterministic appearance and motion descriptors, independently of the LLM judges. Both videos are sampled at $B=16$ uniformly spaced relative positions.

\paragraph{Appearance similarity ($\mathrm{AS}_i$)} is the mean cosine similarity between $\ell_2$-normalized 16-bin RGB histograms at corresponding positions.

\paragraph{State similarity ($\mathrm{SS}_i$)} compares the start, middle, and end frames. Each descriptor concatenates 8-bin per-channel RGB histograms with a 16-bin gradient-magnitude histogram of a $32{\times}32$ grayscale frame, and the cosine similarities are weighted toward the final state:
\begin{equation}
    \mathrm{SS}_i = 0.20\,s_i^{\mathrm{start}} + 0.30\,s_i^{\mathrm{middle}} + 0.50\,s_i^{\mathrm{final}}.
\end{equation}

\paragraph{Temporal alignment ($\mathrm{TA}_i$)} compares motion-intensity curves, the normalized mean absolute grayscale difference between adjacent samples. If both curves vary, $\mathrm{TA}_i=(\rho_i+1)/2$ with Pearson correlation $\rho_i$; if both are constant, $\mathrm{TA}_i=1$; if only one is constant, $\mathrm{TA}_i=\max(0,\,1-20|\mu_i^{\mathrm{gen}}-\mu_i^{\mathrm{gt}}|)$ with mean intensities $\mu_i$. The result is clipped to $[0,1]$.

\medskip
\noindent The final score is
\begin{equation}
    \mathrm{Alignment}_i = 100\left(0.40\,\mathrm{AS}_i + 0.30\,\mathrm{SS}_i + 0.30\,\mathrm{TA}_i\right)\in[0,100].
\end{equation}
Because it compares low-level appearance and motion, Alignment complements rather than replaces the task metrics.

\begin{table}[t]
\centering
\small
\caption{The five VBench dimensions reported in Tables~\ref{tab:main_results} and~\ref{tab:main_results2}, with the predictor used by the official implementation. All are perceptual or temporal quality measures; none assesses procedural correctness.}
\label{tab:vbench_metrics}
\setlength{\tabcolsep}{4pt}
\begin{tabular}{@{}p{0.28\linewidth}p{0.27\linewidth}p{0.39\linewidth}@{}}
\toprule
\textbf{Dimension} & \textbf{Predictor} & \textbf{Measures} \\
\midrule
Aesthetic Quality (AQ) & LAION CLIP aesthetic & Mean frame-level aesthetic score \\
Imaging Quality (IQ) & MUSIQ & Technical image quality; penalizes blur, noise, and distortion \\
Dynamic Degree (DD) & RAFT optical flow & Percentage of videos classified as containing sufficient motion \\
Motion Smoothness (MS) & Frame interpolation & Agreement between observed and interpolated intermediate frames \\
Overall Consistency (OC) & ViCLIP & Video--text semantic consistency with the task description \\
\bottomrule
\end{tabular}
\end{table}

\subsection{VBench Video-Quality Metrics}
\label{app:vbench-evaluation}

We use the official VBench implementation~\citep{zheng2025vbench} for the five dimensions in Table~\ref{tab:vbench_metrics}, reported on a 0--100 scale (higher is better). They measure perceptual and temporal quality, not procedural correctness.

\section{Human Evaluation of Generated Videos}
\label{app:human_model_study}

\paragraph{Setup.}
We conduct a human evaluation with four evaluators, nine models, and 18 task cases, resulting in 162 evaluated videos and 648 individual ratings. For each video, the evaluator first reads the task description and then watches the complete generated video. The evaluator assigns a task-completion score from 1 to 5.

Evaluators are asked to consider three aspects when assigning the score: 
(1) whether the necessary actions are visibly performed in a sensible order,
(2) whether the essential steps are completed, and
(3) whether the final task goal is visibly achieved.
A higher score indicates more successful task completion.

\paragraph{Results.}
For each model, we average its 72 human ratings and report the results in Table~\ref{tab:human_model_study}. WorldGuide achieves the highest average score of 2.750, followed by MiniMax-H3 with 2.542 and LTX-2.5 with 2.472. This human evaluation provides an independent validation of the task-completion results obtained with the automatic video judge.

\section{Additional Results}
\label{app:additional_results}

\subsection{WorldGuide Bench Qualitative Comparisons}
\label{app:additional_qualitative}

\begin{wraptable}{r}{0.4\textwidth}
\centering
\small
\caption{\textbf{Human evaluation of procedural video generation.}
Average task-completion scores on a 1--5 scale. Higher scores indicate better task execution. Model labels follow Table~\ref{tab:main_results}.}
\label{tab:human_model_study}
\begin{tabular}{lr}
\toprule
\textbf{Model} & \textbf{Avg. score / 5} \\
\midrule
Bernini* & 1.583 \\
Cosmos-Predict2.5 & 1.000 \\
HunyuanVideo-1.5 & 1.000 \\
LTX-2.5 & 2.472 \\
MiniMax-H3 & 2.542 \\
PhysAgent* & 1.000 \\
SkyReels & 1.917 \\
TempAct* & 1.444 \\
\midrule
\textbf{WorldGuide} & \textbf{2.750} \\
\bottomrule
\end{tabular}
\vspace{-1em}
\end{wraptable}

\begin{figure}[p]
    \centering
    \includegraphics[width=\linewidth,height=0.82\textheight,keepaspectratio]{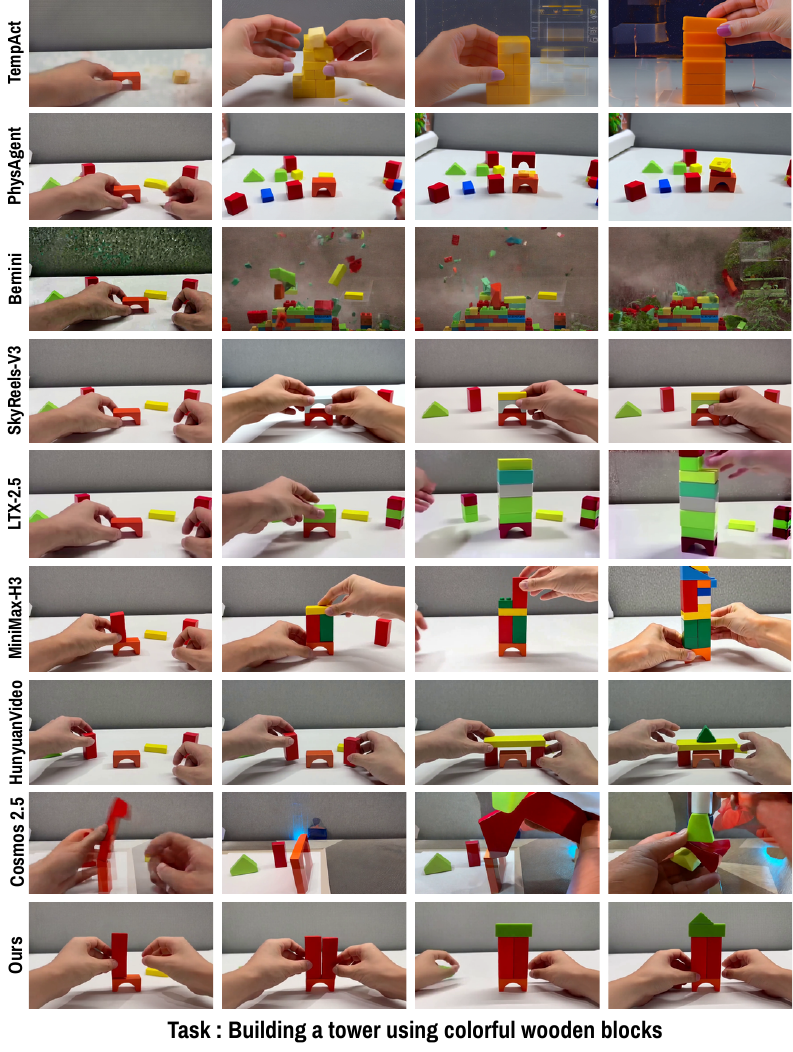}
    \caption{Building a tower from colorful wooden blocks. WorldGuide progressively adds upright pieces and a top to the arch base. TempAct substitutes different blocks, Bernini shows floating pieces and scene changes, and PhysAgent retains several small stacks.}
    \label{app_fig:result1}
\end{figure}

\begin{figure}[p]
    \centering
    \includegraphics[width=\linewidth,height=0.82\textheight,keepaspectratio]{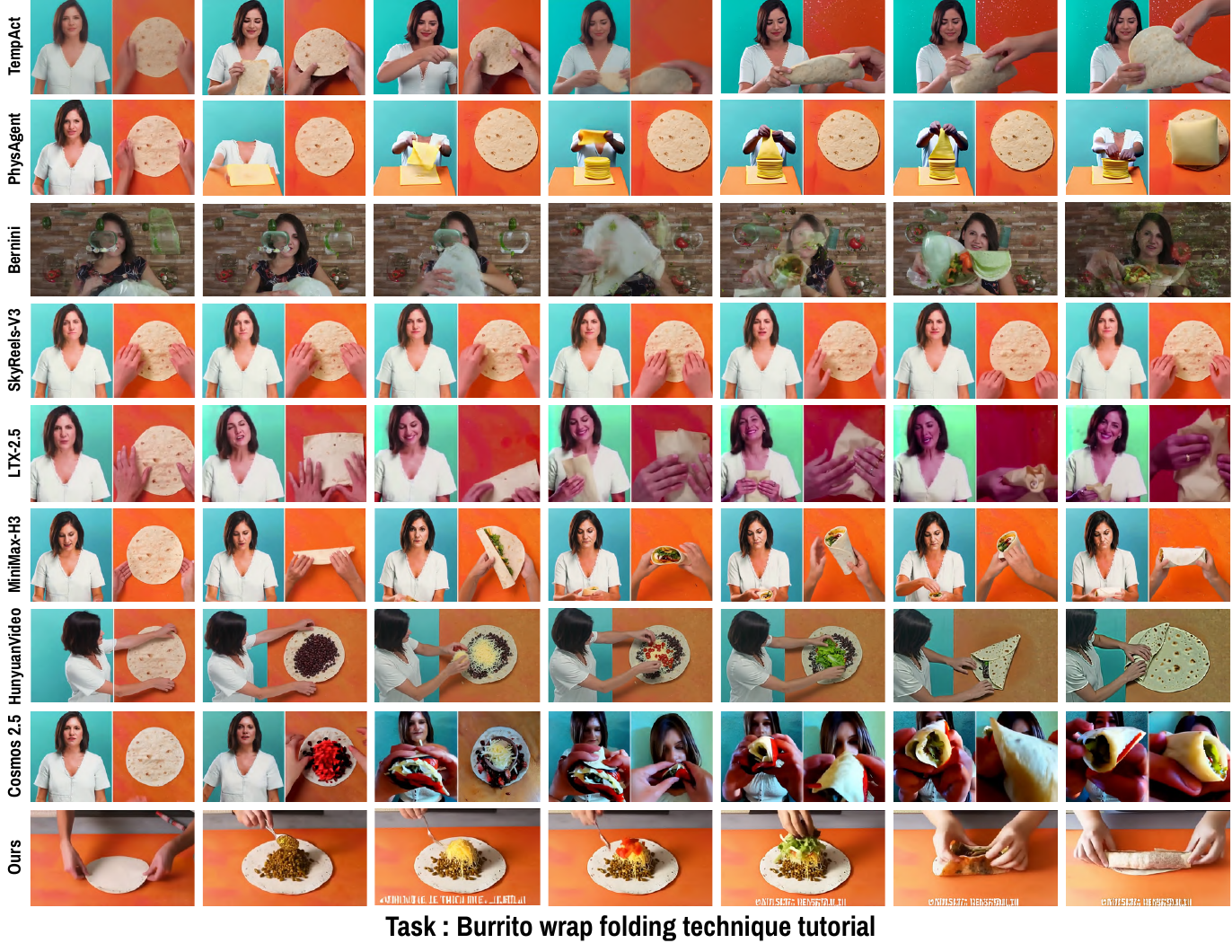}
    \caption{Burrito assembly and folding. WorldGuide adds fillings before folding the tortilla. TempAct folds a tortilla without visible fillings, PhysAgent shifts to stacking round objects, and Bernini replaces the split-screen demonstration with a different scene.}
    \label{app_fig:result2}
\end{figure}

\begin{figure}[p]
    \centering
    \includegraphics[width=\linewidth,height=0.82\textheight,keepaspectratio]{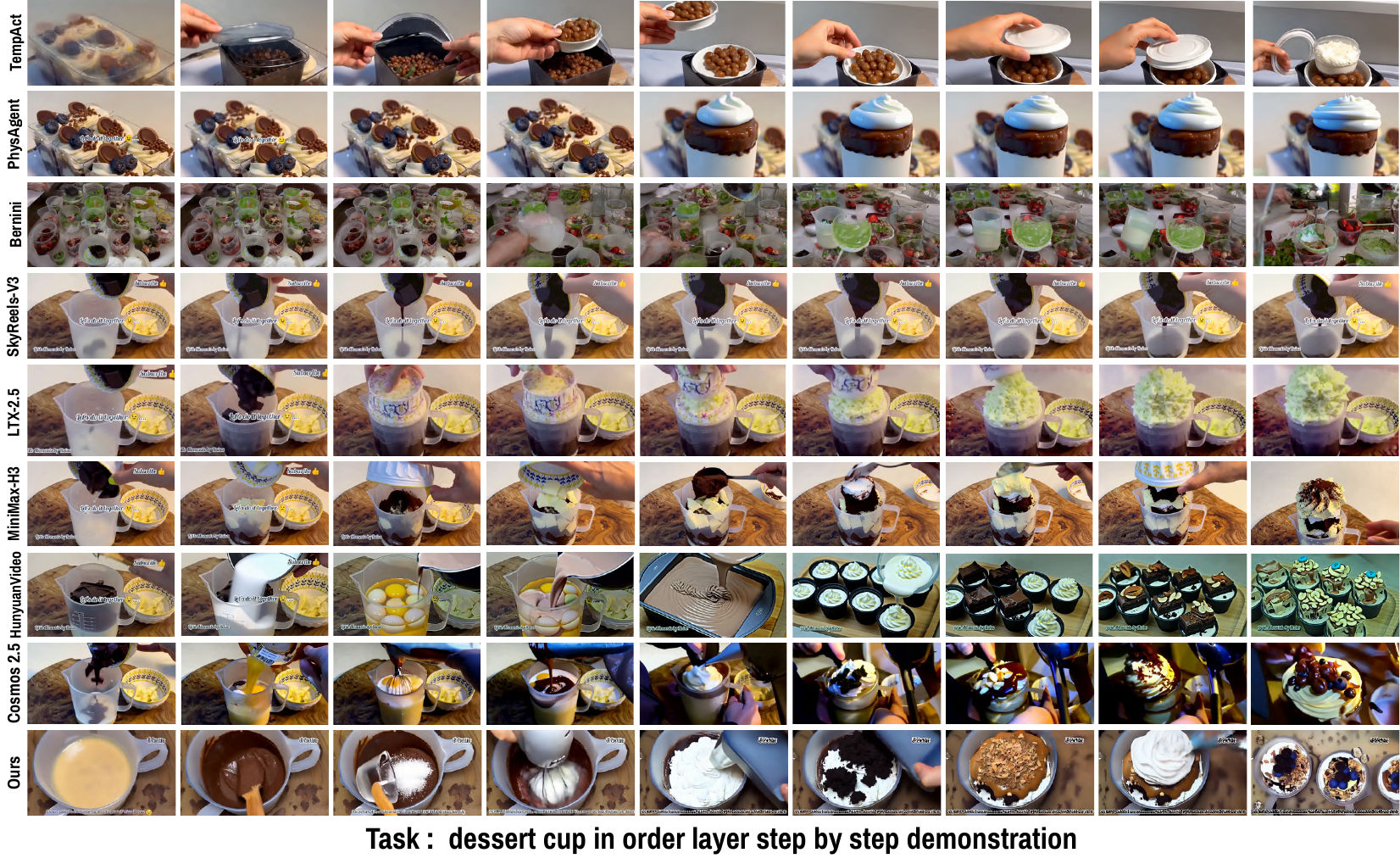}
    \caption{Layered dessert preparation. WorldGuide shows mixing, cream addition, and toppings. HunyuanVideo-1.5 and Cosmos-Predict2.5 also show preparation stages, while Helios develops severe distortion and SpMem changes the vessel and contents.}
    \label{app_fig:result4}
\end{figure}

Figs.~\ref{app_fig:result1}--\ref{app_fig:result4} compare WorldGuide (Ours) with 12 baselines, including Bernini, PhysAgent, and TempAct. The latter three use their own planning--execution loops; the other baselines receive reference action sequences. WorldGuide predicts actions from the goal and generated history. Frames in each row follow temporal order and illustrate action progression and visual continuity; complete rollouts are used for task scoring. Complete generated videos for all compared methods are available on the project page.

The comparisons show distinct limitations. In block construction, TempAct changes the supplied pieces and Bernini introduces floating blocks (Fig.~\ref{app_fig:result1}); in burrito assembly, TempAct folds a tortilla without visible fillings and PhysAgent shifts to stacking round objects (Fig.~\ref{app_fig:result2}). WorldGuide shows successive assembly and preparation steps (Figs.~\ref{app_fig:result1}--\ref{app_fig:result4}); its late visual degradation in folding is shown in Appendix~\ref{app:videocraft_results}.

\subsection{Video-CraftBench Results}
\label{app:videocraft_results}

Table~\ref{tab:main_results2} reports the goal-conditioned results under the protocol of Appendix~\ref{app:eval_protocol}, and Table~\ref{tab:ablation_worldguide} the corresponding memory ablation.

\begin{figure}[p]
    \centering
    \includegraphics[width=\linewidth,height=0.82\textheight,keepaspectratio]{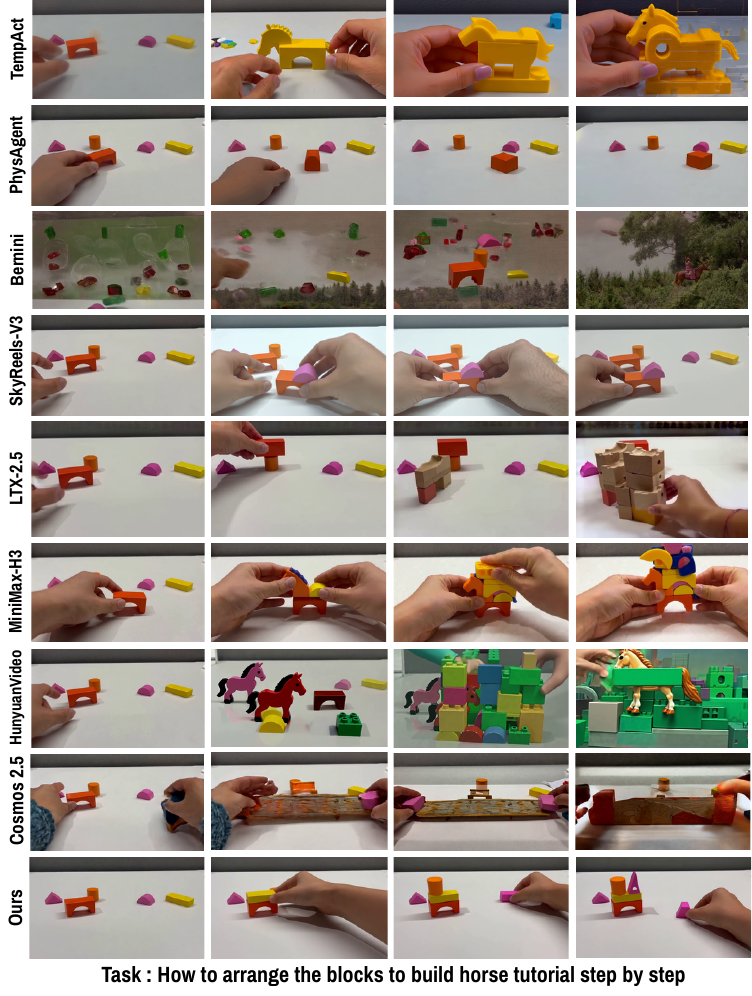}
    \caption{Video-CraftBench: a second block-horse example. WorldGuide adds a yellow bar and orange cylinder to the starting arch. MiniMax-H3 introduces shaped toy pieces, while LTX-2.5 substitutes different blocks.}
    \label{app_fig:videocraft2}
\end{figure}

\begin{figure}[p]
    \centering
    \includegraphics[width=\linewidth,height=0.82\textheight,keepaspectratio]{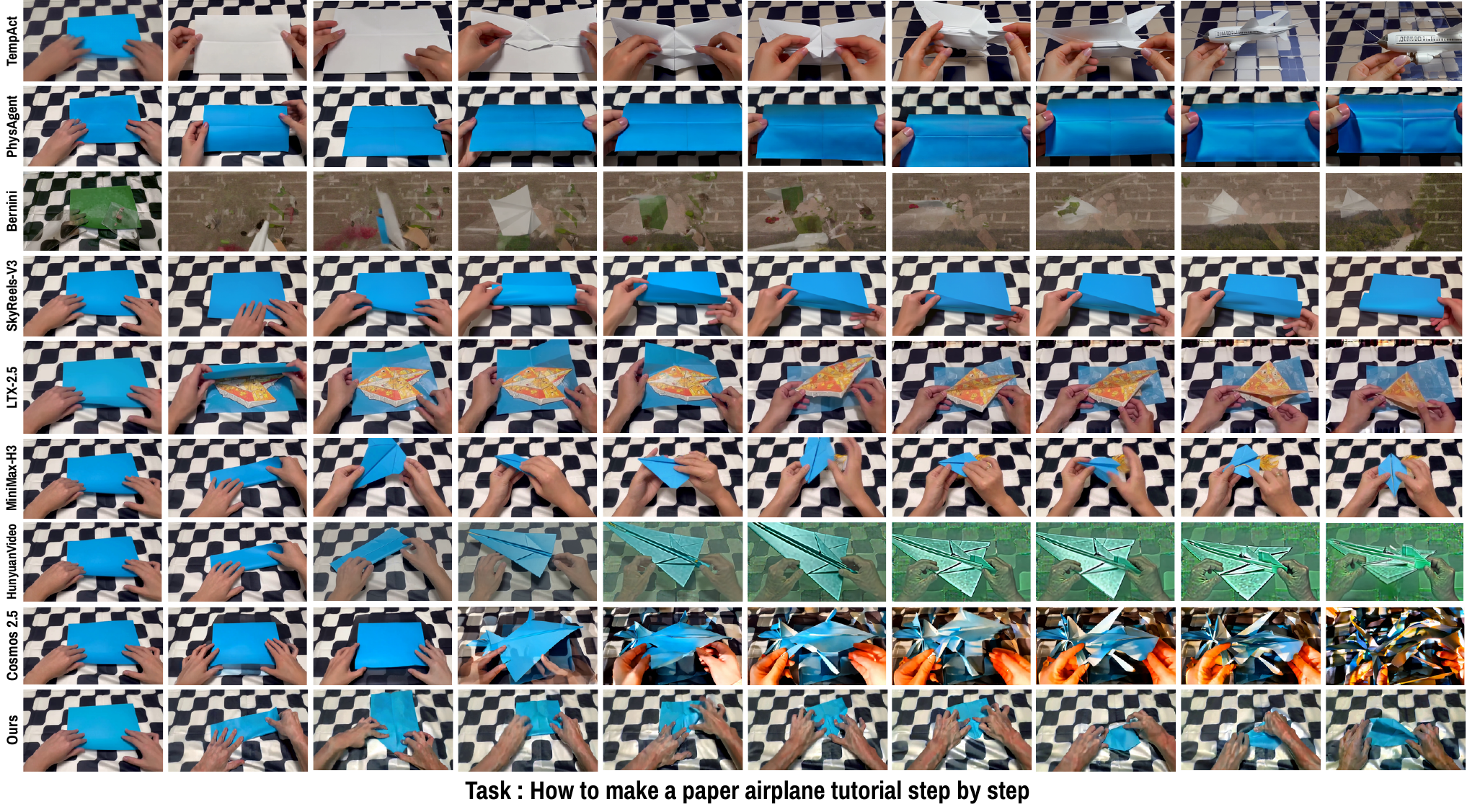}
    \caption{Video-CraftBench: folding a blue paper airplane. WorldGuide progresses through several folds, with visible late texture distortion. PhysAgent repeatedly handles a broad fold, while TempAct changes the blue sheet into a white model aircraft.}
    \label{app_fig:videocraft5}
\end{figure}

\begin{figure}[p]
    \centering
    \includegraphics[width=\linewidth,height=0.82\textheight,keepaspectratio]{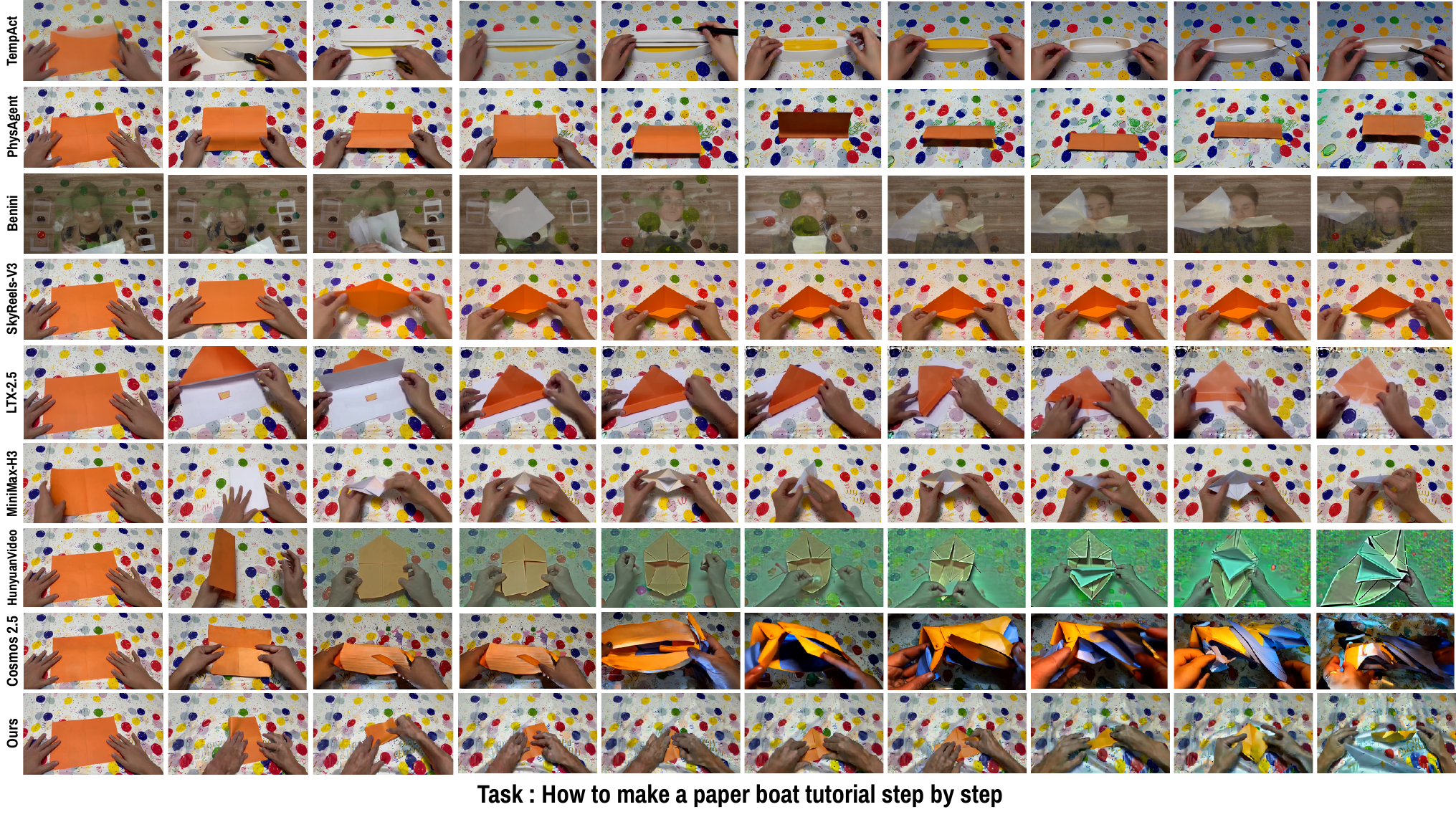}
    \caption{Video-CraftBench: folding an orange paper boat. WorldGuide shows successive folds with late changes in paper color and scene texture. TempAct substitutes a white boat-shaped object, while PhysAgent remains near a rectangular fold.}
    \label{app_fig:videocraft6}
\end{figure}

Figs.~\ref{app_fig:videocraft2}--\ref{app_fig:videocraft6} compare the same 13 models on assembly and folding, with all models receiving an initial image and the task goal, without reference action captions. In the block-horse example, WorldGuide adds a yellow bar and orange cylinder to the starting arch, while MiniMax-H3 introduces shaped toy pieces and LTX-2.5 substitutes different blocks (Fig.~\ref{app_fig:videocraft2}).

Folding remains challenging. TempAct changes the blue sheet into a white model aircraft in the airplane example (Fig.~\ref{app_fig:videocraft5}) and substitutes a white boat-shaped object in the boat example (Fig.~\ref{app_fig:videocraft6}). WorldGuide shows successive folds, but its late paper appearance and scene texture deteriorate in Figs.~\ref{app_fig:videocraft5} and~\ref{app_fig:videocraft6}.


\section{Limitations and Failure Analysis}
\label{app:failure_analysis}

Errors can compound across steps. The ContextPlanner may repeat an action, skip a prerequisite, or stop too early; the Executor may distort geometry or appearance even for a correct instruction. Because each generated clip becomes the next planning observation, replanning can respond to such errors but cannot undo an incorrect transition. The folding examples in Figs.~\ref{app_fig:videocraft5} and~\ref{app_fig:videocraft6} show correct action progression alongside late visual degradation.

The two modules are trained sequentially, without a joint task-completion objective (Appendix~\ref{app:design_rationale}). The bounded memory compresses older frames, which can discard details needed later, and enabling memory also increases repetition (Table~\ref{tab:ablation_worldguide}). Outcome verification, explicit recovery from failed steps, and joint training are natural next steps.

\end{document}